\documentclass[11pt]{article}

\usepackage[final]{acl}

\usepackage{times}
\usepackage{latexsym}
\usepackage[T1]{fontenc}
\usepackage[utf8]{inputenc}
\usepackage{microtype}
\usepackage{inconsolata}
\usepackage{graphicx}

\usepackage{booktabs}
\usepackage{multirow}
\usepackage{amsmath}
\usepackage{amssymb}
\usepackage{mathtools}
\usepackage{amsthm}
\usepackage{wrapfig}
\usepackage{colortbl}
\usepackage{float}
\usepackage{subcaption}
\usepackage{xcolor}
\usepackage[ruled,vlined]{algorithm2e}
\SetKwInOut{KwIn}{Input}
\SetKwInOut{KwOut}{Output}
\SetAlCapSkip{0.5em}

\usepackage[most]{tcolorbox}

\definecolor{primary}{RGB}{62, 136, 177}
\definecolor{light}{RGB}{189, 208, 223}
\definecolor{pink1}{RGB}{234, 228, 242}

\usepackage{tikz}

\title{Beyond FLOPs: Benchmarking Real Inference Acceleration of LLM Pruning under a GEMM-Centric Taxonomy}

\author{
  Haozhe Hu $^{1,2}$ \quad Hao Wu $^{1,2}$ \quad Anhao Zhao $^{2,3}$ \quad Longwei Ding $^2$ \quad Peiran Yin $^{1,2}$ \\
  \textbf{Yunpu Ma}$^4$ \quad \textbf{Xiaoyu Shen} $^2\thanks{Corresponding Author}$ \\
  $^1$Shanghai Jiao Tong University \\
  $^2$Zhejiang Key Laboratory of Industrial Intelligence and Digital Twin, EIT, Ningbo \\
  $^3$Department of Computing, The Hong Kong Polytechnic University\\
  $^4$Munich Center for Machine Learning, LMU Munich\\
  \\\small{ \href{mailto:email@domain}{Sophie17@sjtu.edu.cn; xyshen@eitech.edu.cn}}
}

\begin{document}
\maketitle
\begin{abstract}
Pruning has emerged as a dominant paradigm for accelerating large language model (LLM) inference, spanning a broad spectrum of methods that remove computation across tokens, layers, heads, dimensions, and attention patterns. Despite sharing the same objective, these pruning approaches induce fundamentally different execution behaviors, causing realized speedups to depend heavily on hardware and kernel implementations. Consequently, the practical acceleration benefits of different pruning families remain poorly understood. In this work, we introduce a GEMM-centric taxonomy that reorganizes existing pruning methods according to the logical \textbf{M}, \textbf{N}, and \textbf{K} dimensions of general matrix multiplication (GEMM). Leveraging this abstraction, we build a unified benchmarking framework that enables implementation-consistent comparison across the pruning design space and systematically characterizes the acceleration--quality Pareto frontier. Our results on Llama3.1-8B show that static depth pruning remains the strongest Pareto-optimal baseline and stays closest to its theoretical acceleration upper bound in memory-bounded scenarios. During prefill, the frontier transitions from static depth at low quality loss (0\%--4\%), to dynamic depth at moderate loss (5\%--16\%), and finally to static width pruning at higher loss levels (17\%--26\%). These findings establish the first unified view of the practical limits of pruning-based LLM acceleration and provide guidance for future pruning research.\footnote{Code is available at \url{https://github.com/EIT-NLP/LLM-Pruning/tree/main/PruningInferSim}}

\end{abstract}

\section{Introduction}

\begin{figure}[!t]
  \centering
  \makebox[\linewidth][l]{
    \includegraphics[width=0.9\linewidth,trim=10 10 0 0,clip]{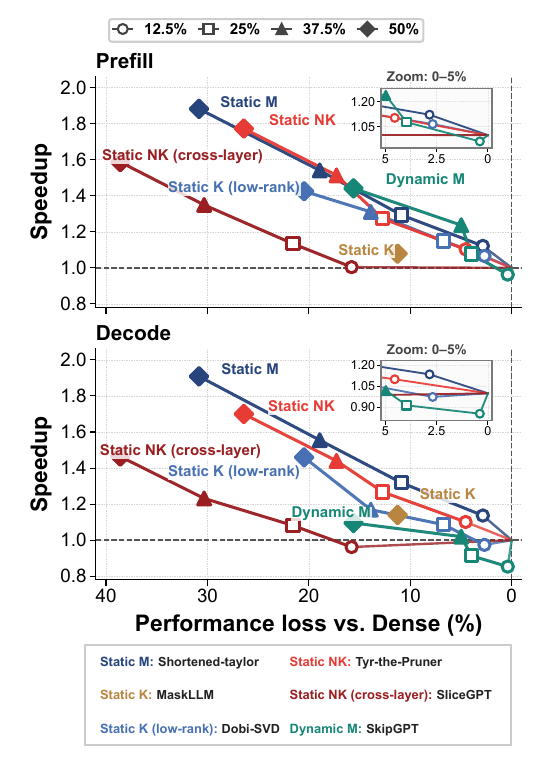}
  }
  \caption{Throughput--quality trade-off of different pruning methods at 12.5\%/25\%/37.5\%/50\% sparsity in Llama3.1-8B, with compared methods listed below.}
  \label{fig:pareto_front}
\end{figure}

The rapid advancement of LLMs has driven progress in reasoning, code generation, scientific discovery, and agentic systems~\cite{zhao2025Survey, hu2025Survey, team2026Kimi}. However, inference efficiency remains a key bottleneck for practical deployment. Acceleration methods span system-level optimizations (e.g., scheduling and kernels~\cite{kwon2023Efficient, zheng2024SGLang,he2026skipopufpgabasedoverlayprocessor}) and model-level compression, among which pruning is widely studied to eliminate redundancy while preserving architectural and deployment compatibility~\cite{zhou2024Survey}, which can naturally extend to Vision Language Models (VLMs) and other multi-modality models~\cite{wu2025HiDrop,wu2026UTPTrack,wu2026Survey,liu2026VICA,tan2026vicostream}. Existing LLM pruning methods span a diverse design space characterized by execution behavior and structural focus. Behaviorally, approaches divide into \emph{static pruning}, which removes a fixed amount of computation, and \emph{dynamic pruning}, which adapts computation per input instance~\cite{jiang2024DLLM,han2025informedroutingllmssmarter,hu2026WIDE}. Structurally, pruning targets various architectural levels, including tokens, layers, and sublayers~\cite{men2025ShortGPT, he2024What,fan2026visualtokensreallyencode}, attention heads~\cite{ma2023LLMPruner}, matrix dimensions~\cite{ashkboos2023SliceGPT}, low-rank approximations~\cite{wang2024SVDLLM}, and attention patterns~\cite{sun2026Efficient}. 

Despite belonging to the broader pruning family, diverse pruning designs are difficult to compare directly. Because these techniques alter a model's shape, sparsity patterns, or execution flow, they directly affect low-level hardware behaviors such as memory access contiguity and kernel scheduling efficiency. As a result, realized speedups become tightly coupled to specific hardware implementations. This dependency severely fragments the evaluation landscape. Since measuring actual latency requires dedicated system engineering for each new method, studies frequently default to theoretical proxies such as FLOPs reduction and single-kernel throughput~\cite{ashkboos2023SliceGPT,fang2024MaskLLM,raposo2024MixtureofDepths,qinsi2024DobiSVD}. In other cases, they evaluate against inconsistent standards by using different model families~\cite{touvron2023llama} or specialized hardware optimizations~\cite{li2025Large}. This leaves a critical gap in understanding \emph{how different pruning patterns translate into practical speedups under a unified and comparable setting}.

Motivated by this gap, we revisit LLM pruning through the lens of GEMM-dominated inference. Regardless of their structural focus, existing pruning methods ultimately reduce computation along one or more of the three GEMM logical axes: M, N, and K. Mapping diverse pruning strategies into this shared operational space abstracts away method-specific complexities and successfully decouples algorithms from bespoke hardware dependencies.

Building on this abstraction, we develop a comprehensive suite that captures both taxonomy-level behaviors and method-specialized implementation details, while providing cross-platform kernel baselines through DSLs such as Triton~\cite{tillet2019Triton} and Tilelang~\cite{wang2025TileLang}. Because any pruning technique can be mapped directly to this shared abstraction, methods can finally be evaluated fairly against an identical baseline. This framework also establishes a plug-and-play foundation for future research, enabling new pruning algorithms to be tested for true end-to-end latency out of the box, while offering a taxonomy-based optimization view that can guide further specialized kernel engineering.

Equipped with this framework, we conduct the first comprehensive cross-design comparison spanning the existing pruning design space. This allows us to systematically benchmark the acceleration potential of each taxonomy in a completely implementation-consistent manner, uncovering the true downstream performance trade-offs of different pruning families for the first time (Figure~\ref{fig:pareto_front}). In summary, our main contributions are as follows:

\noindent (1) \textbf{Taxonomy.} We present a GEMM-centric taxonomy for LLM pruning that unifies existing methods by their logical \textbf{M}, \textbf{N}, and \textbf{K} pruning dimensions, providing an operation-level view of their effect on propagation and implementation.

\noindent (2) \textbf{Framework.} We build a unified inference benchmarking framework with hardware-agnostic baseline acceleration implementations for each taxonomy, enabling controlled comparison of their end-to-end and kernel-level speedup, and further hardware-specific optimizations.

\noindent (3) \textbf{Insights.} Under the same framework and hardware environment, we establish the current Pareto frontier of pruning-based acceleration: static depth remains the strongest speed-oriented baseline, dynamic depth becomes preferable under moderate quality loss, and static width becomes competitive only at higher loss levels.

\begin{figure}[!t]
  \centering
  \includegraphics[width=0.9\linewidth,trim=0 30 0 0,clip]{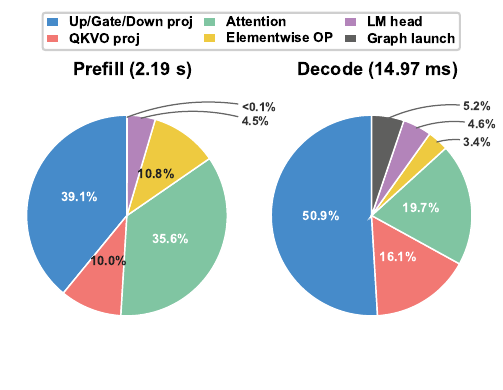}
  \caption{Llama3.1-8B's latency breakdown on each forward step, with batch size 1 and 32,768 context length.}
  \label{fig:sm120_dense_fwd_trace}
  \vspace{-5pt}
\end{figure}

\section{Background and Definitions}
\paragraph{Inference Process of LLM.}
For a standard dense LLM~\cite{grattafiori2024Llama}, we consider the simplest single-step forward pass, where inference cost mainly comes from GEMM computation, element-wise operations (e.g., normalization, residual addition), and the associated CPU-side processing overhead. In practice, the computation in each LLM layer is dominated by nine GEMM operations, which are organized across the attention and feed-forward network (FFN) as follows:
\begin{equation*}
\begin{aligned}
& [Q, K, V] = [X\mathbf{W}_q, X\mathbf{W}_k, X\mathbf{W}_v] \\
& X_{\text{attn.}} = \left(\operatorname{Softmax}(QK^\top/\sqrt{d})V\right)\mathbf{W}_o \\
& X_{\text{FFN}} = \left((X\mathbf{W}_{\text{up}})\odot \sigma(X\mathbf{W}_{\text{gate}})\right)\mathbf{W}_{\text{down}}
\end{aligned}
\end{equation*}
Specifically, the attention layer contains four linear projections (Q, K, V, and O), together with two GEMMs inside the attention computation ($\mathbf{QK}^\top$ and $\mathbf{PV}$), while the FFN layer contains three linear projections (Up, Gate, and Down). Figure~\ref{fig:sm120_dense_fwd_trace} reports the runtime breakdown of these GEMMs and other operations. The breakdown shows that GEMM operations, together with the LM head, account for roughly 90\% of the total inference cost and clearly dominate the overall runtime. This suggests that GEMM-oriented optimization remains a central direction for LLM inference acceleration.

\begin{table*}[!t]
  \centering
  \resizebox{\linewidth}{!}{
    \begin{tabular}{l lc cc cccccc}
      \toprule
      \multirow{2}{*}{\textbf{Categories}} & \multirow{2}{*}{\textbf{Methods}} & \multirow{2}{*}{\textbf{Strategies}} & \multicolumn{2}{c}{\textbf{Raw Dimension}} & \multicolumn{6}{c}{\textbf{MNK-pruning Dimension}}\\
      \cmidrule(lr){4-5} \cmidrule(lr){6-11}
      & & & Attn. & FFN & $\mathbf{Q}$\&$\mathbf{K}$\&$\mathbf{V}$ & $\mathbf{QK}^\top$ & $\mathbf{PV}$ & $\mathbf{O}$ & Up \& Gate & Down\\ 
      \midrule
      \multirow{2}{*}{Static \textbf{M}} & ShortenedLLaMA & depth & \multicolumn{2}{c}{layer} & \bf M & \bf M & \bf M & \bf M & \bf M & \bf M\\
      & BlockPruner & depth & layer & layer & \bf M & \bf M & \bf M & \bf M & \bf M & \bf M\\
      \midrule
      \multirow{4}{*}{Static \textbf{K}} & SVD-LLM$^\dagger$ & low-rank & col & col & \bf K & -- & -- & \bf K & \bf K & \bf K\\
      & Dobi-SVD$^\dagger$ & low-rank & col & col & \bf K & -- & -- & \bf K & \bf K & \bf K\\
      & SparseGPT & semi-structured & col & col & \bf K & -- & -- & \bf K & \bf K & \bf K\\
      & MaskLLM & semi-structured & col & col & \bf K & -- & -- & \bf K & \bf K & \bf K\\
      \midrule
      \multirow{4}{*}{Static \textbf{NK}} & FLAP & width & head & row\&col & \bf N & \bf M & \bf M & \bf K & \bf N & \bf K\\
      & LLMPruner & width & head & row\&col & \bf N & \bf M & \bf M & \bf K & \bf N & \bf K\\
      & Týr-the-Pruner & width & head & row\&col & \bf N & \bf M & \bf M & \bf K & \bf N & \bf K\\
      & SliceGPT$^\ddagger$ & width & row\&col & row\&col & \bf K & -- & -- & \bf N & \bf K & \bf N\\
      \midrule
      \multirow{2}{*}{Dynamic \textbf{M}} & MoD & depth & \multicolumn{2}{c}{layer} & \bf M & \bf M & \bf M & \bf M & \bf M & \bf M\\
      & SkipGPT & depth & layer & layer & \bf M & \bf M & \bf M & \bf M & \bf M & \bf M\\
      \midrule
      \multirow{2}{*}{Dynamic \textbf{NK}} & SeerAttention & sparse attention & key\&value & & -- & \bf N & \bf K & -- & -- & --\\
      & BLASST & sparse attention & value & & -- & -- & \bf K & -- & -- & --\\
      \bottomrule
    \end{tabular}
  }
  \caption{Original pruning strategies, raw dimension, and corresponding pruned \textbf{MNK}-based dimension over each GEMM group. $\dagger$ for static \textbf{K} denotes the low-rank variant, and $\ddagger$ for static \textbf{NK} denotes the cross-layer variant.}
  \label{tbl:dim_corresponding}
  \vspace{-10pt}
\end{table*}

\paragraph{LLM Pruning.}
In the LLM pruning literature, existing approaches are commonly grouped into two categories by pruning dimension: \textbf{depth} and \textbf{width}~\cite{Cheng2024Survey}. \textbf{Depth pruning} removes layers or sublayers, which can be viewed as skipping their computation for all tokens~\cite{kim2024Shortened, men2025ShortGPT, ding2025Pruning, zhong2025BlockPruner,fan2025visipruner}. \textbf{Width pruning}, in contrast, removes neuron groups within weight matrices, such as attention heads (e.g., 128 neurons)~\cite{an2023FLAP,ma2023LLMPruner,li2025TyrthePruner}, entire columns~\cite{ashkboos2023SliceGPT}, finer-grained semi-structured 2:4 groups~\cite{frantar2023SparseGPT,sun2023Simple,fang2024MaskLLM}, and individual neurons~\cite{xia2023FlashLLM}. Beyond zeroing out neurons, the matrix width can also be reduced by low-rank approximation~\cite{wang2024SVDLLM,qinsi2024DobiSVD}, while long-context sparse attention can be regarded as another pruning form that operates over key-value tokens~\cite{gao2025SeerAttention, zhang2025SpargeAttention, yuan2025BLASST}.

Apart from pruning dimension, the dynamics of pruning also affect speed and performance. \textbf{Static pruning} eliminates redundant structures through a one-time calibration phase and loads only the retained neurons during inference~\cite{frantar2023SparseGPT,ma2023LLMPruner}, which reduces memory footprint and requires no modifications to the original implementation for acceleration. \textbf{Dynamic pruning}, conversely, skips computations on-the-fly during inference. For example, token-wise dynamic pruning employs online routing to determine which computations to skip for each token, which often requires customized kernels for real-world batch serving and usually does not directly reduce the memory footprint~\cite{jiang2024DLLM,zhao2025SkipGPT,ding2026llmslrmsrethinkingpruning}. These categories describe pruning granularity and deployment behavior, but they do not reveal which GEMM dimensions are reduced. Since such dimension-level changes directly affect propagation and deployable speedup, we next reinterpret LLM pruning through the \(M\), \(N\), and \(K\) axes of GEMM.
\begin{figure}[!t]
  \centering
  \makebox[\linewidth][l]{
    \includegraphics[width=1.0\linewidth, trim=550 30 550 0, clip]{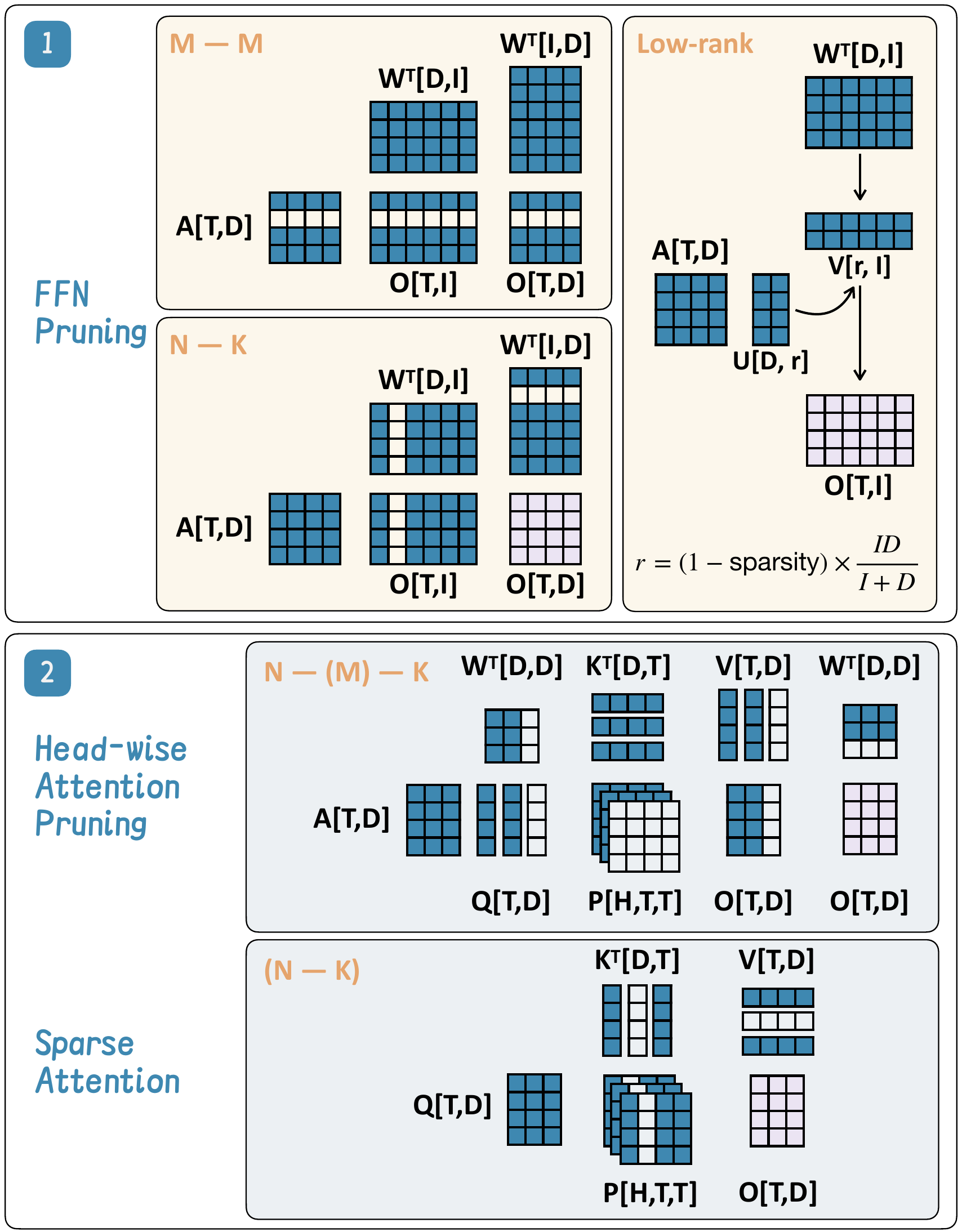}
  }
  \caption{An overview of \textbf{MNK}-dimension pruning and their propagation through attention and FFN layers. The \textbf{empty} blocks denote zero-filled elements that make no contributions to the computation, \textbf{lavender} blocks denote deviations compared to the dense results. The head-wise pruning can also be treated as \textbf{M}-dimension pruning over each head with head dim set to 1, where sparse attention can be treated as pruning over key and value \textbf{N}, \textbf{K} logical dimension, respectively.}
  \label{fig:pruning_example}
\end{figure}

\section{A GEMM-Centric View}
Viewed through GEMM operations, diverse pruning methods exhibit recurring dependency patterns despite their differences in target modules, atomic pruning units, and affected dimensions. In particular, the dimensions pruned in one GEMM can determine how sparsity propagates to subsequent GEMMs within a layer. This motivates a unified GEMM-centric view for characterizing both inter-family differences and intra-family commonalities in implementation behavior.

In general, consider a GEMM formed as $\mathbf{O}=\mathbf{A}\mathbf{W}^\top$ with input tensor $\mathbf{A}\in\mathbb{R}^{\textbf{T}\times\textbf{D}}$, weight $\mathbf{W}\in\mathbb{R}^{\textbf{I}\times\textbf{D}}$, and output tensor $\mathbf{O}\in\mathbb{R}^{\textbf{T}\times\textbf{I}}$. The token dimension \textbf{T}, output feature dimension \textbf{I}, and input feature dimension \textbf{D} correspond to the logical GEMM dimensions \textbf{M}, \textbf{N}, and \textbf{K}, respectively. Although pruning may change the physical shape [\textbf{T, I, D}], this correspondence remains unchanged. Following this mapping, Table~\ref{tbl:dim_corresponding} summarizes representative pruning methods by their original pruning units and corresponding MNK dimensions.

\paragraph{Properties of MNK-dimension Pruning.}
Different pruning dimensions affect not only individual operators, but also the computational flow across an entire layer and even the whole model. For structured zeroing-based pruning, the pruning effect can be summarized in a mask-based manner:
\begin{equation*}
\resizebox{0.9\columnwidth}{!}{$
\begin{aligned}
  \mathbf{M}:\quad
  &\left\{
  \begin{aligned}
  \mathbf{O}
  &=
  (\mathbf{A}\odot \mathcal{M})\mathbf{W}^{\top}
  =
  (\mathbf{A}\mathbf{W}^{\top})\odot \mathcal{M} \\
  &\text{where } \mathcal{M}\in\{0,1\}^{T\times 1}
  \end{aligned}
  \right.
  \\
  \mathbf{N}:\quad
  &\left\{
  \begin{aligned}
  \mathbf{O}
  &=
  \mathbf{A}(\mathbf{W}^{\top}\odot \mathcal{M})
  =
  (\mathbf{A}\mathbf{W}^{\top})\odot \mathcal{M} \\
  &\text{where } \mathcal{M}\in\{0,1\}^{1\times I}
  \end{aligned}
  \right.
  \\
  \mathbf{K}:\quad
  &\left\{
  \begin{aligned}
  \mathbf{O}
  &=
  (\mathbf{A}\odot \mathcal{M})\mathbf{W}^{\top}
  =
  \mathbf{A}(\mathbf{W}\odot \mathcal{M})^{\top} \\
  &\text{where } \mathcal{M}\in\{0,1\}^{1\times D}
  \end{aligned}
  \right.
\end{aligned}
$}
\end{equation*}
Here, each mask is broadcast along its length-1 dimension. A key distinction follows immediately: unlike the \textbf{M} and \textbf{N} dimensions, \textbf{K}-dimension pruning acts on the reduction dimension and therefore has no direct masking equivalent on $\mathbf{O}$.

Now consider a subsequent GEMM that takes $\mathbf{O}$ as input, namely $\mathbf{O}\mathbf{\hat{W}}^{\top}$ with $\mathbf{\hat{W}}\in\mathbb{R}^{\mathbf{D}\times\mathbf{I}}$. For \textbf{M} and \textbf{N} dimension pruning, the masked output becomes $((\mathbf{A}\mathbf{W}^{\top})\odot\mathcal{M})\mathbf{\hat{W}}^{\top}$. In the \textbf{N}-pruning case, the mask applied to the out-feature of the first GEMM becomes a mask on the in-feature dimension of the next GEMM; that is, it reappears as \textbf{K}-dimension pruning in the subsequent operator. This yields the propagation behavior shown in Figure~\ref{fig:pruning_example}: \emph{\textbf{M}-dimension pruning persists across consecutive GEMMs; \textbf{N}-dimension pruning propagates as \textbf{K}-dimension pruning in the next GEMM, forming an \textbf{NK} pattern; \textbf{K}-dimension pruning exhibits no quantifiable propagation.}

Low-rank pruning differs from the zeroing-based cases above in that it compresses the weight matrix through low-rank factorization rather than explicit masking. Although the output \textbf{N}-dimension remains unchanged, the reduced latent dimension lowers computation without structural propagation across subsequent GEMMs. We therefore treat it as a variant of static \textbf{K}.

Overall, this GEMM-centric view shows why nominal sparsity is an unreliable predictor of realized acceleration: pruning dimensions differ in the computations they remove and the implementation constraints they impose, motivating our unified implementation and benchmarking framework.

\section{Inference Implementation}
After establishing the pruning taxonomy, we study how each sparsity pattern translates into practical acceleration. For generality and fair comparison, we build a taxonomy-level acceleration framework rather than method-specific implementations. The framework provides a unified operator-replacement interface, reusable optimization patterns for each pruning family, and specialized kernels written in high-level DSLs such as Triton~\cite{tillet2019Triton} and TileLang~\cite{wang2025TileLang}.

\paragraph{Static K.}
For static \textbf{K}, the implementation is straightforward since neither method causes structural propagation. In both cases, we replace the original operator with a weight-level branch, implemented via low-rank factorization or Sparse Matrix Multiplication (SpMM). Moreover, the naive PyTorch implementation of semi-structured sparsity incurs substantial CPU overhead, so we replace it with a JIT interface that directly invokes cuSPARSELt, reducing initialization latency by 94.4\% (726$\mu$s $\rightarrow$ 40$\mu$s) with improved architecture compatibility. We further run auto-tuning for each logical \textbf{M} size to select the best SpMM configuration (Appendix~\ref{sec:Static K}).

\paragraph{Static M and NK.}
For static methods with structural propagation, the key implementation issue is to maintain dimension consistency along dependent operators, rather than to redesign kernels. Static \textbf{M} pruning removes entire layers or sublayers, whereas static \textbf{NK} pruning couples consecutive modules through \textbf{N}-to-\textbf{K} propagation. We thus implement them with a unified wrapper that rewires each dependent module group jointly. Nevertheless, not all static \textbf{NK} share the same path. The cross-layer static \textbf{NK}, which is represented by SliceGPT~\cite{ashkboos2023SliceGPT}, begins at token embedding and attention output projection, extends to the LM head, but it also introduces an extra residual projection and excluding propagation through attention computations (Table~\ref{tbl:dim_corresponding}). Accordingly, we apply pruning in a coarse-to-fine order and treat each coupled group as a single unit. The overall pipeline is given as Algorithm~\ref{alg:propagated_pruning}. Here, the GEMMs outside the attention are divided into four groups by data dependencies (Table~\ref{tbl:dim_corresponding}), where the GEMMs in each group share the same pruning dimension.
\begin{algorithm}[!t]
  \caption{Propagated Structure Pruning}
  \label{alg:propagated_pruning}
  \KwIn{model $\mathcal{M}$, pruning config $\mathcal{P}$}
  \KwOut{pruned model $\mathcal{M}'$}

  \ForEach{layer $\ell$ in $\mathcal{M}$}{
      \ForEach{sublayer $s \in \{\text{Attention}, \text{MLP}\}$}{
          \If{\textsc{ShouldPruneM}$(s_\ell, \mathcal{P})$}{
              \textsc{PruneM}$(\mathcal{M}, s_\ell)$\;
          }
      }
  }

  $\mathcal{C} \leftarrow \textsc{GetGroups}(\mathcal{M}, \mathcal{P})$\;
  \If{\textsc{CrossLayer}$(\mathcal{P})$}{
      $\mathcal{C} \leftarrow [embedding,\ \mathcal{C},\ lm\ head]$\;
  }

  \ForEach{$(src, dst)$ in $\mathcal{C}$}{
      \If{\textsc{Prunable}$(src, dst, \mathcal{P})$}{
          $\mathcal{I} \leftarrow \textsc{GetIndices}(src, dst, \mathcal{P})$\;
          \textsc{PruneNK}$(\mathcal{M}, src, dst, \mathcal{I})$\;
          \If{\textsc{CrossLayer}$(\mathcal{P})$ and \textsc{HasResidual}$(src, dst)$}{
              \textsc{AddProjection}$(src, \mathcal{M})$\;
          }
      }
  }
  \Return $\mathcal{M}$\;
\end{algorithm}

\begin{figure}[!t]
  \centering
  \includegraphics[width=1.0\linewidth, trim=400 300 400 300, clip]{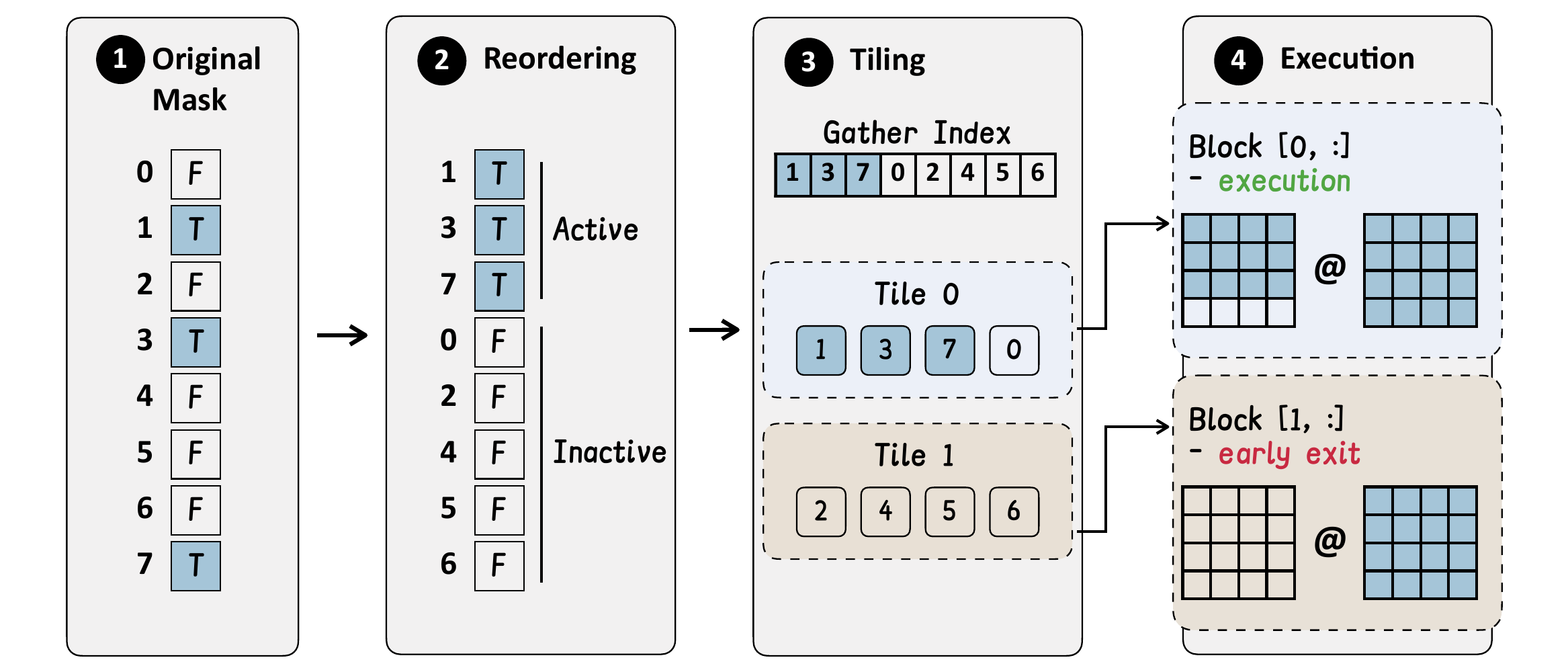}
  \caption{The execution pipeline of dynamic \textbf{M}, with the number of tokens set to 8, and M-axis tile size to 4, blocks that contain any active tokens will be executed.}
  \label{fig:dynamic_m_example}
  \vspace{-5pt}
\end{figure}

\paragraph{Dynamic M and NK.}
For dynamic pruning, we convert online sparsity into speedup through mask preprocessing and intra-kernel skipping, avoiding method-specific pipeline abstraction. Dynamic \textbf{M} and dynamic \textbf{NK} thus share a tile-based execution principle: the former derives sparse masks from token-wise routing, while the latter derives them from KV-page masking. For dynamic \textbf{M}, a naive implementation performs mask-to-indices conversion and gather--scatter fusion with indices $\mathcal{I}$\footnote{Mask-to-indices conversion triggers dynamic memory allocation at runtime, which breaks the static graph.}:
\begin{equation*}
\resizebox{0.95\columnwidth}{!}{$
\begin{aligned}
  \mathcal{K}(X, \mathbf{W}, \mathcal{I})
  &=
  \mathrm{Scatter}_{\mathcal{I}}\!\left(
  \mathrm{OP}\!\left(
  \mathrm{Gather}_{\mathcal{I}}(X), \mathbf{W}
  \right)
  \right)
\end{aligned}
$}
\end{equation*}
Thus an additional mask-reordering preprocessing scheme is critical for such tile-based intra-kernel skipping (Figure~\ref{fig:dynamic_m_example}). Beyond these hardware-agnostic optimizations, several decoding-specific optimizations are also integrated to ensure a fair comparison. Details are provided in Appendix~\ref{sec:Implementation Details}.

\begin{table*}[!t]
    \centering
    \aboverulesep=0ex
    \belowrulesep=0ex
    \renewcommand{\arraystretch}{1.2}
    \resizebox{1.0\linewidth}{!}{
        \begin{tabular}{l|cccc|cccc}
            \toprule
            \multirow{2}{*}{\textbf{Taxonomy}} & \multicolumn{4}{c}{\textbf{25\% sparsity}} & \multicolumn{4}{c}{\textbf{50\% sparsity}}\\
            \cmidrule(lr){2-9}
            & \textbf{WikiText2} & \textbf{Acc. Gap \%} & \textbf{Prefill Speedup} & \textbf{Decode Speedup} & \textbf{WikiText2} & \textbf{Acc. Gap \%} & \textbf{Prefill Speedup} & \textbf{Decode Speedup}\\
            \midrule
            Dense & 7.54 & 0.00 & 1.00x [1.00x, 1.00x] & 1.00x [1.00x, 1.00x] & 7.54 & 0.00 & 1.00x [1.00x, 1.00x] & 1.00x [1.00x, 1.00x]\\
            \midrule
            Static \textbf{M} & 15.52 & 10.59 & \textbf{1.29x} [1.28x, 1.34x] & \textbf{1.32x} [1.31x, 1.34x] & 33.93 & 30.28 & \textbf{1.88x} [1.85x, 1.92x] & \textbf{1.91x} [1.89x, 1.94x]\\
            Static \textbf{K}$^\dagger$ & 10.14 & \underline{6.52} & 1.15x [1.11x, 1.19x] & 1.09x [0.97x, 1.23x] & 15.44 & 19.83 & 1.43x [1.31x, 1.53x] & 1.46x [1.38x, 1.57x]\\
            Static \textbf{K} & -- & -- & -- & -- & \underline{11.45} & \textbf{10.67} & 1.08x [1.03x, 1.20x] & 1.14x [1.08x, 1.26x]\\
            Static \textbf{NK} & 12.46 & 12.63 & \underline{1.27x} [1.24x, 1.35x] & \underline{1.27x} [1.23x, 1.39x] & 19.80 & 26.10 & \underline{1.77x} [1.62x, 1.84x] & \underline{1.70x} [1.62x, 1.84x]\\
            Static \textbf{NK}$^\ddagger$ & 38.96 & 21.16 & 1.13x [1.12x, 1.16x] & 1.08x [1.02x, 1.17x] & 68.71 & 38.36 & 1.59x [1.37x, 1.76x] & 1.46x [1.26x, 1.59x]\\
            Dynamic \textbf{M} & \underline{9.25} & \textbf{3.94} & 1.08x [0.98x, 1.12x] & 0.91x [0.75x, 1.05x] & 13.14 & \underline{15.44} & 1.44x [1.13x, 1.56x] & 1.10x [0.76x, 1.44x]\\
            Dynamic \textbf{NK}$^*$ & \textbf{7.72} & 33.40 & 1.02x [1.00x, 1.06x] & 1.02x [1.00x, 1.04x] & \textbf{7.76} & 34.45 & 1.05x [0.98x, 1.20x] & 1.04x [1.01x, 1.10x]\\
            \bottomrule
        \end{tabular}
    }
    \caption{Overview of WikiText2 perplexity, average accuracy gap, and speedup over the dense model under different pruning taxonomies. Throughput speedups are reported in \emph{mean [min, max]}. The 1st and 2nd results are highlighted in \textbf{bold} and \underline{underline}, respectively. $*$ for dynamic \textbf{NK} denotes to force 25\%/50\% sparsity in all tasks.}
    \label{tbl:overview_table}
\end{table*}

\section{Experiments}~\label{sec:Experiments}
This section studies the practical deployment behavior of different pruning methods beyond nominal sparsity. We organize the evaluation around three questions: 
\emph{
(1) which pruning taxonomies form the throughput--quality Pareto frontier; 
(2) how realized prefill and decode speedups differ under a unified setup; and 
(3) where the gap between theoretical acceleration bounds and realized speedup comes from.
}
Finally, we distill these findings into taxonomy-level implications, outlining the practical role, current bottlenecks, and remaining optimization headroom of each pruning family.

\subsection{Experimental Setup}
We select Llama3.1-8B~\cite{grattafiori2024Llama} and Qwen3-4B~\cite{yang2025Qwen3} as the baseline, then apply a unified LoRA (Low-rank Adaptation)~\cite{hu2021LoRA} fine-tuning recipe to each method after calibration. Unless otherwise noted, training runs for 5,000 steps with batch size 16, sequence length 2,048, and LoRA rank 16 on a subset of RedPajama~\cite{weber2024redpajama}. For dynamic \textbf{M}, we extend training to 10,000 steps for convergence. For static \textbf{K}, continued tuning is infeasible because its semi-structured sparsity pattern does not support vanilla LoRA merging. At each sparsity level, we select the method with the best LoRA performance as the representative of its taxonomy. More details about the baseline selection are provided in Appendix~\ref{sec:Implementation Details}.

For throughput simulation, the hardware is set to RTX Pro 6000 Blackwell. All profiling experiments are integrated with CUDA graph to minimize CPU-side overhead. For each pruning strategy, we randomly sample their atomic units to simulate based on target sparsity budget. Throughput is computed as the total number of processed tokens divided by the elapsed wall-clock time:
\begin{align*}
  \text{Token/s} = \frac{\textbf{B}\times\textbf{T}_q}{\textbf{TTFT}\ \text{or}\ \textbf{TPOT}}
\end{align*}
where \textbf{B} is the batch size, $\textbf{T}_q$ is the number of tokens for each input sequence, TTFT (Time-to-First-Token) and TPOT (Time-Per-Output-Token) denote the execution time. Additionally, all width-pruned dimensions are aligned to multiples of 16 to ensure maximum speedup (Table~\ref{tbl:static_width_alignment} and~\ref{tbl:triton_gemm_align_test}).
\begin{figure}[t]
  \centering
  \includegraphics[width=0.9\linewidth, trim=10 0 0 0, clip]{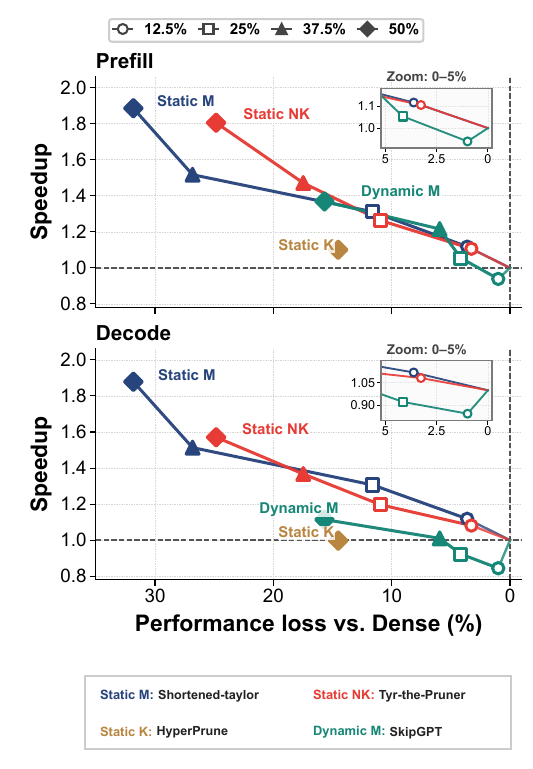}
  \caption{Throughput--quality trade-off of different pruning methods at 12.5\%/25\%/37.5\%/50\% sparsity in Qwen3-4B, with compared methods listed below.}
  \label{fig:pareto_front_qwen3}
  \vspace{-15pt}
\end{figure}
\begin{figure*}[!t]
  \centering
  \makebox[\linewidth][l]{
    \includegraphics[width=1.0\linewidth,trim=20 10 10 0,clip]{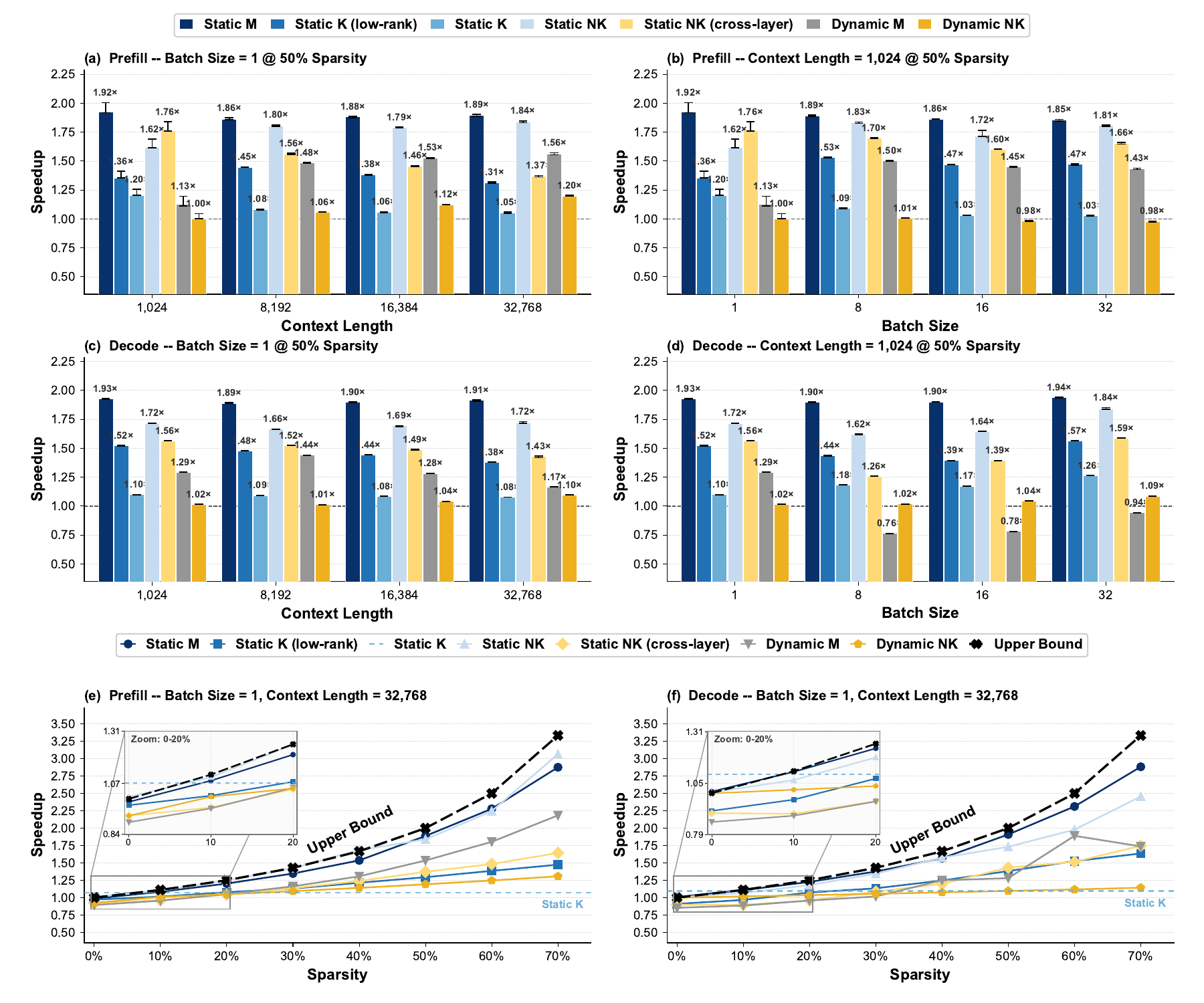}
  }
  \caption{(a-d) Prefill and decode speedup on Llama3.1-8B under 50\% sparsity. (e-f) Speedup over different sparsity, and their gap with the theoretical upper bound. Context in decoding denotes the KV cache length.}
  \label{fig:sm120_50_fuse}
  \vspace{-10pt}
\end{figure*}

\subsection{Throughput--Quality Trade-off}

\begin{tcolorbox}
\small
\textbf{Finding 1:} The Pareto frontier shifts with pruning taxonomy and quality budget: static \textbf{M} dominates low loss, dynamic \textbf{M} leads moderate loss, and static \textbf{NK} becomes competitive at higher loss.
\end{tcolorbox}
The results reveal that pruning does not yield a single universal throughput--quality frontier; instead, the frontier shifts with pruning taxonomy and quality budget. As shown in Figure~\ref{fig:pareto_front}, Table~\ref{tbl:overview_table}, and Table~\ref{tbl:performance_lora}, static \textbf{M} provides the strongest trade-off at the low-loss end, achieving a 1.12x speedup across evaluated scenarios with only 2.76\% performance loss under 12.5\% sparsity. As the quality budget relaxes, dynamic \textbf{M} is most competitive in the moderate-loss regime, improving prefill speedup to 1.24x$\sim$1.44x at roughly 5\%$\sim$16\% loss while remaining more quality-efficient than higher-sparsity static baselines. At larger loss levels, static \textbf{NK} becomes the strongest width-pruning frontier method, reaching 1.51x speedup at 16.97\% loss and 1.77x at 26.10\% loss. By contrast, the two static \textbf{K} variants, static \textbf{NK} (cross-layer), and dynamic \textbf{NK} remain off the main frontier due to weaker realized speedup, poorer quality retention, or both. The attention-only dynamic \textbf{NK} preserves quality mainly in long-context settings, such as WikiText2. Overall, these results show that practical acceleration is determined not by sparsity alone, but by pruning taxonomy and how effectively its induced structure translates into realized throughput.

To assess whether the Pareto frontier remains stable across models, we further evaluate Qwen3-4B, focusing on the three pruning taxonomies that reach the frontier on Llama3.1-8B: static \textbf{M}, dynamic \textbf{M}, and static \textbf{NK}. As shown in Figure~\ref{fig:pareto_front_qwen3}, the prefill frontier transitions from static \textbf{M} at 0$\sim$5\% performance loss, to dynamic \textbf{M} at 6$\sim$10\%, and then to static M (11$\sim$13\%) and static \textbf{NK} (13$\sim$24\%), similar to the pattern observed on Llama3.1-8B. For decoding, static \textbf{M} dominates the 0$\sim$18\% loss range, whereas static \textbf{NK} leads at 19$\sim$24\%. Thus, despite several model-dependent shifts, the overall pattern remains consistent across models. Within the models and methods evaluated, each pruning taxonomy therefore occupies a distinct regime in which it offers the best throughput--quality trade-off.

\subsection{Prefill and Decode Throughput}

\begin{tcolorbox}
\small
\textbf{Finding 2:} Static \textbf{M} provides the most stable speedup across prefill and decode, while width and dynamic pruning show substantially different realized gains even under the same sparsity budget.
\end{tcolorbox}
As shown in Figures~\ref{fig:sm120_50_fuse} and~\ref{fig:sm120_50_appendix}, a consistent pattern across both prefill and decode is that static methods provide more stable acceleration than dynamic ones. Among them, static \textbf{M} remains the strongest baseline throughout, showing that \textbf{M}-axis pruning translates most directly into realized end-to-end gains. At 50\% sparsity in the prefill stage, the average speedups of static \textbf{K} (low-rank), static \textbf{NK}, static \textbf{NK} (cross-layer), and dynamic \textbf{M} are only comparable to those achieved by static \textbf{M} at 34\%, 48\%, 39\%, and 38\% sparsity, respectively. The same pattern holds in decoding, where static \textbf{K} (low-rank), static \textbf{NK}, static \textbf{NK} (cross-layer), and dynamic \textbf{M} at 50\% sparsity match only the decoding gains of static \textbf{M} at 34\%, 45\%, 37\%, and 23\% sparsity. Width pruning therefore still trails depth pruning by a clear margin even under the same sparsity budget, and this gap is not uniform across width families: static \textbf{NK} is consistently the strongest, whereas static \textbf{K} and static \textbf{NK} (cross-layer) realize substantially smaller gains, especially in attention-dominated long-context scenarios. These results suggest that the main divide is not simply static versus dynamic or depth versus width, but how the pruning pattern is organized and how effectively it removes executable computation in the end-to-end inference pipeline.
\begin{figure*}[!t]
    \centering
    \makebox[\linewidth][l]{
    \includegraphics[width=1.0\linewidth,trim=40 20 0 0,clip]{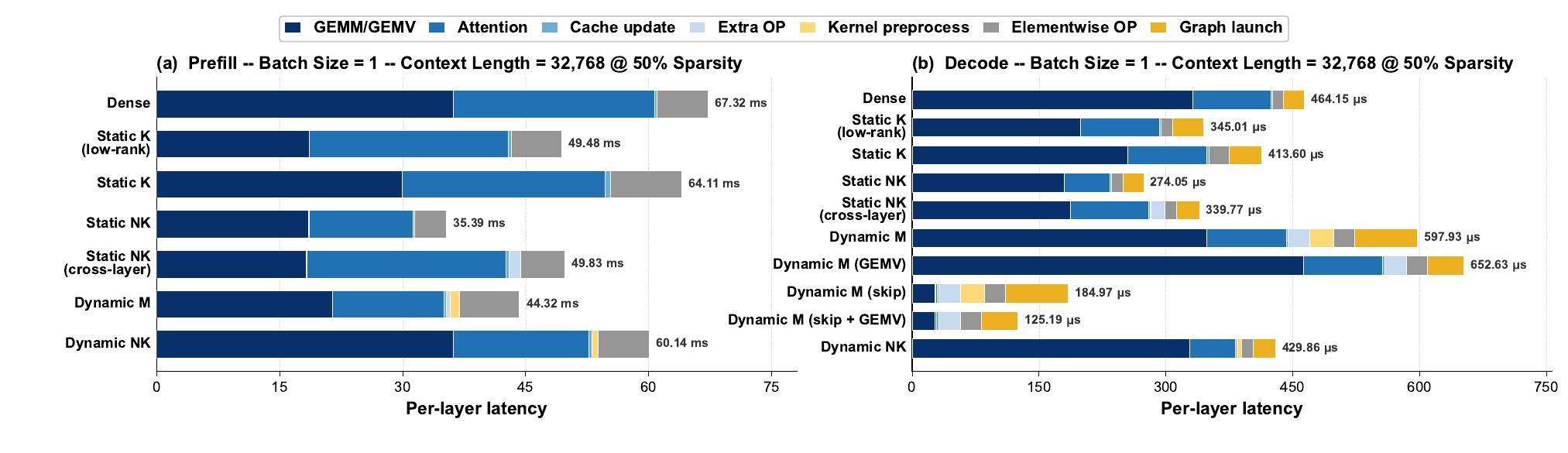}
    }
    \caption{Breakdown of wall-clock time for all operations in one layer. The \emph{extra operation} includes overheads apart from the vanilla GEMM/attention/element-wise kernels (e.g., residual projection and routing), where \emph{kernel preprocess} includes overheads before the core GEMM/attention kernel execution (e.g., mask reordering).}
    \label{fig:sm120_ttft_tpot_trace_breakdown}
    \vspace{-10pt}
\end{figure*}

\subsection{Latency Breakdown}
\begin{tcolorbox}
\small
\textbf{Finding 3:} The gap between theoretical and realized acceleration comes from inefficiency in the GEMMs and taxonomy-specific non-GEMM overhead, which becomes especially critical during decoding.
\end{tcolorbox}
To further analyze how different pruning strategies affect the LLM forward trace, we measure the latency breakdown of core GEMM and remaining non-GEMM operations. Figure~\ref{fig:sm120_ttft_tpot_trace_breakdown} shows that the gap between theoretical and realized acceleration cannot be explained by pruned GEMM throughput alone. In particular, dynamic \textbf{M} and static \textbf{K} both incur substantial auxiliary overheads beyond the main compute kernels. Their non-GEMM overhead increases by 42.4\% and 40.8\% in prefill, and by 61.5\% and 287.2\% in decode, respectively, directly offsetting part of their nominal compute reduction\footnote{Sparse tensor cores only support \textbf{A}\textbf{B}$^\top$ with \textbf{A} in 2:4 sparse format, which requires a (\textbf{B}\textbf{A}$^\top$)$^\top$ execution form.}. Static \textbf{NK} (cross-layer) is similarly affected by its extra residual projection path, while static \textbf{K} (low-rank) also pays additional launch overhead from factorizing one GEMM into two smaller ones. These two sources increase decoding non-GEMM overhead by 48.5\% and 30.9\%, respectively. Meanwhile, the core pruned operators themselves still differ in realized efficiency: dynamic \textbf{M} and static \textbf{K} remain noticeably below their apparent kernel-level headroom, static \textbf{K} (low-rank) reaches only about 1.66$\times$ GEMM speedup in decoding, and dynamic \textbf{NK} still trails static \textbf{NK} in attention throughput under the same sparsity budget (Table~\ref{tbl:attn_prefill_tflops_by_seqlen}). 

Overall, dynamic \textbf{M} and static \textbf{K} remain the two methods furthest from their apparent upper bounds, as both are limited by non-GEMM overheads and GEMM optimization. By contrast, static \textbf{K} (low-rank) and dynamic \textbf{NK} are constrained more by the efficiency of their pruned GEMM operation. These results show that the deployable acceleration of a pruning method depends not only on how much GEMM work it removes, but also on the extra non-GEMM cost it introduces along the pipeline.

\subsection{Taxonomy-Level Implications}
The above results reveal distinct roles and bottlenecks across pruning taxonomies:

\indent \textbf{Static \textbf{M} is the current throughput anchor.} Its coarse-grained pruning pattern largely explains why it remains closest to the speedup bound and delivers stable gains across serving scenarios. The same rigidity, however, also constrains its quality at higher sparsity, especially beyond 37.5\%.

\indent \textbf{Dynamic \textbf{M} is a promising quality-aware direction.} Its token-level routing adaptively maintains model quality well, while this dynamism also introduces challenges for acceleration, especially for decoding's non-GEMM overhead.

\indent \textbf{Static \textbf{NK} reaches the width-pruning frontier so far.} Its finer-grained neuron pruning yields a more favorable quality profile than static \textbf{M} at high sparsity, while the vast pruning search space prevents current methods from fully exploiting this advantage, especially in the low-sparsity regime.

\indent \textbf{Other width-pruning methods remain quality-preserving but not frontier-ready.} Static \textbf{K} generally retain quality well, while static \textbf{NK} (cross-layer) further covers the LM head. In practice, however, their limitations in non-GEMM overhead and GEMM-side optimization are non-negligible.

\indent \textbf{Dynamic \textbf{NK} is currently a long-context-specialized direction.} It mainly captures attention-side sparsity and therefore becomes attractive only when the context is sufficiently long.

\section{Conclusion}
This work presents a GEMM-centric view of LLM pruning that organizes diverse pruning methods through the logical \textbf{M}, \textbf{N}, and \textbf{K} dimensions and their propagation behavior across Transformer computation. Based on this view, we build a unified and reusable benchmarking framework, while evaluating representative pruning taxonomies under fixed sparsity budgets from the perspectives of downstream quality, realized prefill/decode throughput, and kernel-level latency breakdown.

Our results show that nominal sparsity alone is a weak predictor of deployable acceleration. In practice, realized speedup is jointly determined by pruning taxonomy, structural propagation, operator coverage, and system overheads introduced during execution. More broadly, these findings suggest that the next Pareto frontier will likely come not from pushing a single taxonomy in isolation, but from hybrid designs that combine a quality-aware branch with structured and static backbone.

\section*{Limitations}
Our study focuses on taxonomy-level inference behavior rather than exhaustive deployment coverage. The main experiments are conducted on Llama3.1-8B with RTX Pro6000 (sm120), with additional validation on Qwen3-14B and A800 (sm80) in the appendix. Although the trends are consistent across these settings, several limitations remain.

First, we do not evaluate Mixture-of-Experts (MoE) models or other accelerator platforms such as data-center Blackwell (sm100). These settings may require different execution pipelines, including warp-specialized kernels, and may shift the relative advantages of dynamic \textbf{M} and dynamic \textbf{NK}. Second, our Triton and TileLang implementations are designed as portable baselines and may underperform hand-written CUDA kernels or hardware-specific implementations using features such as Tensor Memory Accelerator (TMA) and async tensor core instructions (e.g., \texttt{tcgen05.mma}). Third, we benchmark representative recent methods for each taxonomy rather than exhaustively covering all pruning variants. Therefore, the reported Pareto frontier should be interpreted as the current frontier under our selected methods, models, and benchmark settings, rather than a definitive upper envelope over all possible pruning methods. Finally, our measurements focus on single-step prefill/decode latency and kernel-level behavior, rather than full serving workloads in production inference frameworks such as SGLang. As a result, our findings provide deployment-relevant evidence at the model-step and kernel levels, while system-level scheduling, distributed serving, and production integration remain important directions for future work.

\bibliography{ref}

\newpage
\appendix

\section{The Use of Large Language Models}
We employed large language models (LLMs) solely as general-purpose writing assistants for language refinement, including improving clarity, grammar, and style. Importantly, no LLM was involved in research ideation, methodological design, analysis, or result interpretation; the role of the LLM was limited to linguistic polishing. All substantive contributions originated from the authors. This ensured that the scientific content remained entirely authored by the researchers, while benefiting from improved academic writing quality.

\section{Implementation Details}\label{sec:Implementation Details}
\paragraph{Static K.}\label{sec:Static K}
Semi-structured static \textbf{K} pruning applies a 2:4 mask to the weight matrices, compressing them to 50\% along the \textbf{K} dimension, while obtaining metadata to describe their original coordinates and then invoking sparse tensor core for computation, the bit count of metadata typically equals to the number of elements in the original matrix (2-bit per non-zero element). We followed PyTorch's official cuSPARSELt integration, using SGLang's TVM-FFI\footnote{\url{https://github.com/apache/tvm-ffi}} JIT framework to directly integrate the cuSPARSELt library and bypass the CPU-side overhead in the pytorch implementation. An explicit cache is applied to automatically tune different matrix shapes and maximize reuse of cuSPARSELt's descriptors, thereby significantly reducing kernel preprocessing overhead and CPU-side costs during decoding. For the prefill phase, we used PyTorch's native interface for SpMM and kernel tuning. Notably, in the A800 tests, we found both implementations failed to properly trigger cuSPARSELt's split-k mode, which affect the performance of semi-structured sparsity during decoding on sm80 architecture.

\paragraph{Static K (low-rank).}\label{sec:Static N}
The low-rank variant is conceptually similar to semi-structured, which directly applies the weight-by-weight pruning without propagation. Given a fixed sparsity budget, we first determine the target rank for each matrix, and then monkey-patch the original weight and GEMM path with a low-rank factorization implemented as two consecutive matrix multiplications. Because this transformation reduces the effective intermediate dimension and further increases the number of GEMM launches, static \textbf{K} (low-rank) inevitably falls short of the expected per-GEMM speedup and incurs additional graph launch overhead.

\paragraph{Static M, Static NK.}\label{sec:Static M, Static NK}
We process propagated static structured pruning in a unified order: layer-wise pruning, attention/FFN sublayer pruning, followed by row-column pruning. For depth pruning, we replace the \texttt{forward} function of the corresponding layers with identity mappings. For width pruning, each consecutive module pair is treated as the minimum processing unit, to handle \textbf{N}-dimension pruning in the first module and \textbf{K}-dimension pruning for the second, while simultaneously managing associated element-wise operations (e.g., normalization and residuals). When propagating \textbf{NK} pruning to attention, we enforce head-level granularity: when the number of pruned query heads reaches the head group size, the corresponding key and value heads will also be removed as well.
\begin{algorithm}[!ht]
  \caption{Dynamic \textbf{M}'s GEMM kernel}
  \label{alg:gemm_dynamic_m}
  \KwIn{input matrix $\mathbf{A}\in\mathbb{R}^{M\times K}$, weight matrix $\mathbf{B}\in\mathbb{R}^{N\times K}$, routing mask $\mathbf{m}\in\{0,1\}^{M}$}
  \KwOut{output matrix $\mathbf{D}\in\mathbb{R}^{M\times N}$}

  $(\tilde{\mathbf{m}}, \mathbf{idx}) \leftarrow \textsc{SortDescending}(\mathbf{m})$\;
  Initialize $\mathbf{D}\leftarrow \mathbf{0}$\;

  \ForEach{tile $(b_m, b_n)$ in the output grid}{
      $\mathbf{m}^{(b_m)} \leftarrow \tilde{\mathbf{m}}[b_m\!\cdot\!B_M : (b_m+1)\!\cdot\!B_M]$\;
      \If{$\textsc{All}(\mathbf{m}^{(b_m)}) = 0$}{
          continue\;
      }

      $\mathbf{r} \leftarrow \mathbf{idx}[b_m\!\cdot\!B_M : (b_m+1)\!\cdot\!B_M]$\;
      $\mathbf{M}_{row} \leftarrow (\mathbf{m}^{(b_m)} = 1)$\;
      $\mathbf{C} \leftarrow \mathbf{0}^{B_M \times B_N}$\;

      \For{$k = 0$ \KwTo $\lceil K / B_K \rceil - 1$}{
          $\mathbf{A}_{tile} \leftarrow \textsc{GatherRows}\big(\mathbf{A}, \mathbf{r}, k\big)$\;
          $\mathbf{B}_{tile} \leftarrow \textsc{LoadTile}\big(\mathbf{B}, b_n, k\big)$\;
          $\mathbf{C} \leftarrow \textsc{MMA}(\mathbf{A}_{tile}, \mathbf{B}_{tile}, \mathbf{C})$\;
      }
      \tcp{Element-wise operation}
      \textsc{ApplyEpilogueFuse}$(\mathbf{C})$\;

      \textsc{ScatterRows}$\big(\mathbf{D}, \mathbf{r}, \mathbf{C}, \mathbf{M}_{row}\big)$\;
  }

  \Return $\mathbf{D}$\;
\end{algorithm}

\paragraph{Dynamic M.}\label{sec:Dynamic M}
For dynamic \textbf{M}, we implement the corresponding Flash Attention and GEMM kernels under Triton with autotune enabled. Following the approach in Figure~\ref{fig:dynamic_m_example}, we reorder the routing mask to ensure tile-level processing for both skipped and executed blocks, thereby preserving the design of the original dense kernel as much as possible. Specifically, before the kernel enters the pipeline, it will perform a block-level reduction on routing mask to detect if there are any active tokens in the current block. Additionally, for decoding-stage optimization, the flash decoding, split-k GEMM, and GEMV dispatch are also integrated into the attention and GEMM kernels, which can be automatically triggered when the number of blocks is less than GPU's streaming multiprocessors (SMs), with a heuristic algorithm to minimize the waste of SMs. The online skipping schedule can be summarized as Algorithm~\ref{alg:gemm_dynamic_m}. Additionally, since the key-value projection only occupies a small fraction of the total computation, applying a dynamic skip kernel to these two operations would negligible reduce the overall throughput, so we kept the original dense GEMM for these two operations.

\paragraph{Dynamic NK.}\label{sec:Dynamic NK}
For block-sparse sparse attention in dynamic \textbf{NK}, the overall implementation is similar to the attention part of dynamic \textbf{M}, with the main difference lying in kernel preprocessing, which requires to compute an approximate attention map for $\mathbf{QK}^{\top}$ and derives a fixed skipping matrix. The vanilla attention pipeline only needs to read the mask for the current tile at the first step to determine whether the tile should be skipped. Following the original SeerAttention~\cite{gao2025SeerAttention,gao2025Sparsea}, we also implement the kernel with Tilelang, combined with flash decoding and GQA packing for better decoding performance. Meanwhile, since the original SeerAttention updates the compressed KV cache block-by-block (e.g., every 64 generated tokens), the averaged update overhead per step is less than 1\%, so only the gating computation is considered during profiling.

\paragraph{The Selection of DSL.}
As discussed above, we use Triton and Tilelang to implement dynamic \textbf{M} and dynamic \textbf{NK}, respectively, both are tile-based Python DSLs. Triton is built entirely around block-level pointer programming, with the block as its smallest control granularity, while providing a relatively unified abstraction over load, store, and compute operations through an LLVM-based backend. In contrast, Tilelang uses TVM as its backend and supports both block-level execution and finer-grained thread-level operations. It also exposes hardware-specific interfaces, such as shared memory, and makes different load, store, and compute primitives more explicit. Overall, Tilelang outperforms Triton in both CPU-side overhead and kernel throughput, which is especially evident in the SeerAttention implementation (523.64 TFLOPS vs. 501.74 TFLOPS). However, it still has limitations in predicate operations and pipeline fusion. In particular, during the decoding stage of SeerAttention, enabling flash decoding does not support multi-stage pipelining, where the dynamic \textbf{M} kernels can only be implemented with Triton due to the incomplete optimization for bulk gather and scatter memory-access patterns.

\paragraph{Training Settings.}
For static \textbf{M}, static \textbf{K} (low-rank), static \textbf{NK}, static \textbf{NK} (cross-layer), and dynamic \textbf{M}, we train representative methods within each taxonomy from scratch and report the best-performing one as the taxonomy representative in downstream evaluation. All methods follow the original calibration strategy and training dataset. Because the LoRA fine-tuning dataset can substantially affect downstream performance, while several methods do not include an additional LoRA stage, we standardize LoRA fine-tuning on \texttt{RedPajama-Data-1T-Sample-subset850000}\footnote{\url{https://huggingface.co/datasets/konwoo/RedPajama-Data-1T-Sample-subset850000}}, using learning rate 2e-4, LoRA dropout 0.1, rank 16, and alpha 32. For Dobi-SVD in the static \textbf{K} (low-rank) category, we follow SVD-LLM~\cite{wang2024SVDLLM} to train $\mathbf{U}$ first and $\mathbf{V}$ second, allocating half of the total budget (2,500 steps) to each stage. For static \textbf{K} and dynamic \textbf{NK} with high training overhead, we directly adopt the released Llama3.1-8B checkpoints. In particular, SeerAttention~\cite{gao2025SeerAttention} only releases a Llama3.1-8B-Instruct checkpoint. Since the dense instruct and base models perform nearly identically on our evaluation tasks, we directly reuse its trained attention gate for reference.

\paragraph{Evaluation Settings.}
We use the lm-evaluation-harness~\cite{eval-harness}\footnote{\url{https://github.com/EleutherAI/lm-evaluation-harness}} to conduct all downstream evaluations under a unified setup. The evaluation suite includes the commonly used benchmarks: WikiText2~\cite{wikitext2}, ARC-e, ARC-c~\cite{arc}, BoolQ~\cite{boolq}, WinoGrande~\cite{winogrande}, PIQA~\cite{piqa}, OpenBookQA~\cite{openbookqa}, and HellaSwag~\cite{hellaswag}. All tasks are evaluated in the zero-shot setting with a maximum context length of 4,096. For WikiText2, we report word perplexity (PPL); for BoolQ and WinoGrande, we report accuracy (acc.); and for the remaining tasks, we report normalized accuracy (acc. norm). The overall quality loss and average accuracy metrics for each model only include the aforementioned 7 classification tasks, excluding WikiText2.

As for the baseline in each taxonomy, since exhaustively evaluating every pruning strategy is infeasible, we thus introduce a three-stage screening procedure:
\begin{enumerate}
  \item Collect recent methods and their cited baselines that release code or checkpoints and can be mapped to our wrapper.
  \item Evaluate all candidates at 25\% and 50\% sparsity under the same pruning and LoRA adaptation protocol.
  \item Retain the best candidate in each taxonomy for the remaining sparsity levels.
\end{enumerate}
Following this procedure, we evaluate Shortened-ppl, Shortened-taylor~\cite{kim2024Shortened}, CoopPruner~\cite{ding2025Pruning}, BlockPruner~\cite{zhong2025BlockPruner}, and ReplaceMe~\cite{shopkhoev2025ReplaceMe} for static \textbf{M}; MaskLLM~\cite{fang2024MaskLLM}, HyperPrune~\cite{sun2025Learning}, and Dobi-SVD~\cite{qinsi2024DobiSVD} for static \textbf{K}; FLAP~\cite{an2023FLAP}, T\'yr-the-Pruner~\cite{li2025TyrthePruner}, LoRAP~\cite{li2024LoRAP}, and SliceGPT~\cite{ashkboos2023SliceGPT} for static \textbf{NK}; MoD~\cite{raposo2024MixtureofDepths}, DLLM~\cite{jiang2024DLLM}, and SkipGPT~\cite{zhao2025SkipGPT} for dynamic \textbf{M}; and SeerAttention~\cite{gao2025SeerAttention} for dynamic \textbf{NK}. MaskLLM provides only a Llama3.1-8B checkpoint, so we use HyperPrune for Qwen3-4B. SeerAttention likewise provides only a Llama3.1-8B checkpoint, while training Dobi-SVD incurs substantial computational cost; we therefore evaluate both methods only on Llama3.1-8B.

\paragraph{Profiling Settings.}
We build our unified benchmark framework for pruning methods following the overall design of ELANA~\cite{chiang2025ELANA}, the main different is that it will call a unified pruning interface for all models before profiling, then relies on Triton's built-in benchmarking utilities to flush the L2 cache, run repeated measurements, collect wall-clock times, and generate final reports. The CUDA graph interface is based on native PyTorch APIs, we record an additional forward trace under the given CUDA stream after warmup, and then replay the trace to record the wall-clock time in each step. To support CUDA graph execution and emulate the paged KV-cache update cost, a static-length dummy KV cache is allocated to replace the vanilla dynamic KV cache, thus the KV cache updating in each step is a directly in-place writing operation, avoiding the overhead of online memory allocation.

During the model pruning, the mask and indices are strictly generated by sorting and top-$k$ selection to ensure a consistent distribution across pruning modules. For end-to-end throughput evaluation, we reinitialize each evaluated model three times to sample different pruning distributions and report the mean value over all results (i.e., 3 times resample with 50 times repeating per run). We perform 10 warm-up iterations before each set of 50 sample iterations. For kernel-level latency, we use the PyTorch profiler to sample 10 iterations and report the GPU-side execution time in middle layer (e.g., layer 16) per iteration. For the result in Table~\ref{tbl:static_width_alignment}, we additionally perturb the target pruning ratio assigned to each module, producing non-uniform pruning ratios and unaligned dimensions. In all other experiments, we prune the attention and FFN modules with a unified sparsity ratio, avoiding confounding effects from different sparsity definitions, e.g., the fraction of removed dimensions for static \textbf{NK} and the fraction of skipped tokens for dynamic \textbf{M}.

All the profiling results are run with CUDA 12.8, PyTorch 2.9.1, and Transformers 4.57.1. Semi-structured sparsity uses cuSPARSELt 0.9.0, integrated through the sglang's JIT framework (0.5.8.post1). Custom kernels are implemented with Triton 3.6.0 and Tilelang 0.1.8 for full optimization. All the experiments on RTX Pro6000 are under bf16 precision, where the A800 runs in fp16 precision because sm80 do not support \texttt{atomic add} with bf16 precision. The evaluated models are Llama3.1-8B\footnote{\url{https://huggingface.co/meta-llama/Llama-3.1-8B}} and Qwen3-14B\footnote{\url{https://huggingface.co/Qwen/Qwen3-14B}}.

\section{Additional Experiments}~\label{sec:Additional Experiments}
\begin{figure}[!t]
  \centering
  \makebox[\linewidth][l]{
  \includegraphics[width=1.0\linewidth]{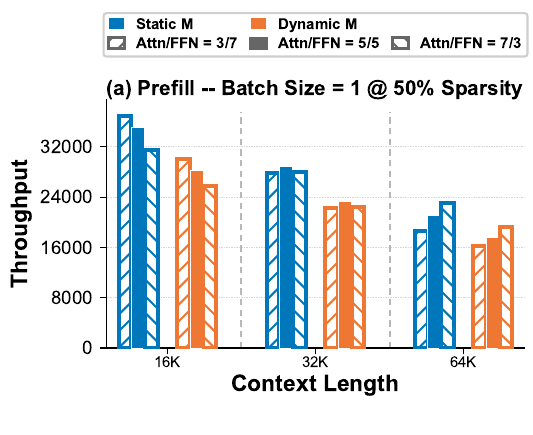}
  }
  \caption{Prefill throughput with 50\% sparsity and different pruning component distribution (Attention/FFN).}
  \label{fig:ttft_diff_distribution_sm120}
\end{figure}

\paragraph{Influence of Sparsity Distribution.}
For layer pruning, the distribution of sparsity across attention and FFN modules can affect the overall speedup at a given sparsity level, for Llama3.1-8B, the theoretical balance where attention and FFN have equal computational cost occurs at $T=4D$ (sequence length $T=16384$, hidden size $D=4096$). When non-GEMM overheads are taken into account, Figure~\ref{fig:ttft_diff_distribution_sm120} and Figure~\ref{fig:sm120_ttft_tpot_trace_breakdown} show that this balance point shifts to around 32k, where attention and FFN consume comparable wall-clock time. Since current pruning methods typically apply higher sparsity to attention than FFN layers~\cite{zhao2025SkipGPT}, a long-context scenario can achieve better acceleration effects under the provided sparsity.

\begin{figure}[htbp]
  \centering
  \makebox[\linewidth][l]{
    \includegraphics[width=1.0\linewidth,trim=0 10 0 0, clip]{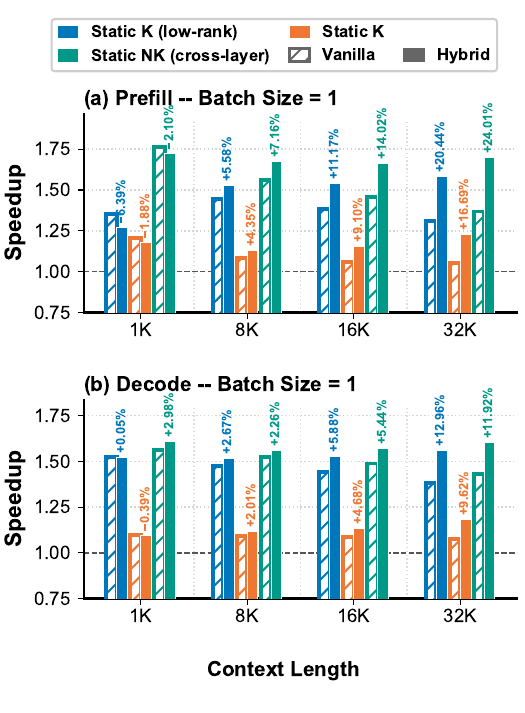}
  }
  \caption{Prefill \& decode speedup on vanilla static \textbf{K} (low-rank)/static \textbf{K}/ static \textbf{NK} (cross-layer) and their variants combined with SeerAttention-style sparse attention.}
  \label{fig:hybrid_ttft_tpot}
\end{figure}

\paragraph{Attention-side acceleration for static K and NK (cross-layer).}
As shown in Figure~\ref{fig:sm120_ttft_tpot_trace_breakdown}, the vanilla static \textbf{K} (low-rank), static \textbf{K}, and static \textbf{NK} (cross-layer) only prune the main linear modules per layer, which causes their suboptimal speedup as sequence length increases. Meanwhile, based on Table~\ref{tbl:dim_corresponding}, these methods naturally eliminate the propagation to attention computation, making them compatible with existing sparse attention techniques. Therefore, we can conveniently combine these approaches to obtain a hybrid width pruning method that also provide benefits under long-context scenarios. As demonstrated in Figure~\ref{fig:hybrid_ttft_tpot}, during prefill and decoding stages at 32k sequence length, both hybrid methods achieve approximately 10\%$\sim$20\% average speedup compared to their original version, respectively, with these gains continuing to increase for longer sequences. While dynamic \textbf{NK} enables static \textbf{NK} (cross-layer) to outperform dynamic \textbf{M} in long-context scenarios, it provides only marginal improvement for static \textbf{K} pruning, which is bounded by its acceleration potential on linear layers.

\paragraph{Influence of Dimension Alignment.}
For width pruning, the modification of parameter matrices shape also impacts the execution efficiency of GEMM operation. To quantify this effect, we evaluate different alignment strategies for the same target sparsity. As results in Table~\ref{tbl:static_width_alignment}~and~\ref{tbl:triton_gemm_align_test}, it is clear that an unaligned layout can cause up to 35\% loss of speedup. Since only 2/7 of GEMM in static \textbf{NK} prunes the leading dimension, it retains a more robust performance than static \textbf{NK} (cross-layer) with 5/7 of pruned GEMM in leading dimension, and static \textbf{K} (low-rank), with 7/14 of the low-rank GEMM expose their rank size dimension $r$ to the logical \textbf{K} dimension.

To further analyze the sensitivity of structured width pruning to dimension alignment, we benchmark the kernel-level GEMM performance over different alignment strategies. As results in Table~\ref{tbl:triton_gemm_align_test}, the trend of standalone GEMM is broadly consistent with Table~\ref{tbl:static_width_alignment}. The non-leading \textbf{N} dimension imposes relatively weak alignment constraints, whereas the \textbf{K} dimension becomes increasingly sensitive to alignment at lower precision. In fp8, when the alignment is smaller than 16 (16$\times$1 bytes), throughput reaches only 11\% of the baselines, but recovers to 70\% once the alignment is a multiple of 16. Since the alignment requirements vary across the thread layout of memory access and compute instructions such as \texttt{mma.sync} under different precision, these results should be treated only as a rough trend. In practical width pruning, the pruning granularity should be at least aligned to (vectorized memory access width / data width) under the target precision, in order to maximize speedup.

\paragraph{Pruning over N vs. Pruning over K.}
To compare the differences between two width-pruning axis at the kernel-level, we further measure their speedups over the dense baseline across different sparsity levels and weight shapes, with results in Table~\ref{tbl:static_nk_vs_static_kn_in_mlp_prefill} and~\ref{tbl:static_nk_vs_static_kn_in_mlp_decode}. During prefill, pruning along the \textbf{N} axis leads to uneven speedups because reducing \textbf{N} also changes the total number of tiles, with the \textbf{M} tiling fixed, speedup appears only when the total number of GPU execution waves decreases. By contrast, pruning along \textbf{K} directly reduces the loop count within each block, lowering per-block runtime and yielding more stable speedups. However, although \textbf{N}-axis pruning in the prefill stage can only beat \textbf{K}-axis pruning in the up and gate projections with larger \textbf{N} dimensions, standard static \textbf{NK} still retains an advantage in linear projection operations due to the 53.84\% proportion of these two GEMM's operation counts in each layer of Llama3.1-8B.

When turns to the batch-size-1 decoding, the \textbf{M} tiling is fixed to 1, forcing both schemes to activate split-k to increase the block count to match the number of SMs, therefore have the similar wave counts, and pruning along \textbf{N} or \textbf{K} can both change the split-k strategy and the runtime of each wave, resulting in a smoother speedups for normal static \textbf{NK} pruning than static \textbf{NK} (cross-layer). In this setting, pruning along \textbf{K} exposes the prologue and epilogue latency of the per-block pipeline under split-k, whereas pruning along \textbf{N} preserves the original loop count along \textbf{K}. As a result, \textbf{N}-dimension pruning achieves better decoding speedups.
\begin{table}[!t]
  \centering
  \caption{Ranking of averaged speedup in Llama3.1-8B across different hardware (RTX Pro6000 / A800 / H20), with S1--S4 corresponded to settings in Figure~\ref{fig:sm120_50_fuse}, i.e., prefill vary context length / batch size, and decode vary context length / batch size.}
  \resizebox{\linewidth}{!}{
    \begin{tabular}{l cccc}
      \toprule
      Taxonomic & S1 & S2 & S3 & S4\\
      \midrule
      Static M & 1/1/1 & 1/1/1 & 1/1/1 & 1/1/1\\
      Static K & 6/6/6 & 6/6/6 & 6/7/6 & 5/5/6\\
      Static K (low-rank) & 5/4/5 & 4/4/4 & 4/4/5 & 3/4/4\\
      Static NK & 2/2/2 & 2/2/2 & 2/2/2 & 2/2/3\\
      Static NK (cross-layer) & 3/3/4 & 3/3/3 & 3/3/3 & 4/3/2\\
      Dynamic M & 4/5/3 & 5/5/5 & 5/5/4 & 7/7/5\\
      Dynamic NK & 7/7/- & 7/7/- & 7/6/- & 6/6/-\\
      \bottomrule
    \end{tabular}
  }
  \label{tbl:stablility_analysis_hardware}
\end{table}

\begin{table}[!t]
  \centering
  \caption{Ranking of averaged speedup in RTX Pro6000 across different model (Llama3.1-8B / Qwen3-14B / Qwen3-4B). Dynamic NK (Tilelang) encounters a compilation error on H20.}
  \resizebox{\linewidth}{!}{
    \begin{tabular}{l cccc}
      \toprule
      Taxonomic & S1 & S2 & S3 & S4\\
      \midrule
      Static M & 1/1/1 & 1/1/1 & 1/1/1 & 1/1/1\\
      Static K & 6/7/6 & 6/6/6 & 6/6/6 & 5/5/5\\
      Static K (low-rank) & 5/4/5 & 4/4/4 & 4/4/5 & 3/4/4\\
      Static NK & 2/2/2 & 2/2/2 & 2/2/2 & 2/2/2\\
      Static NK (cross-layer) & 3/3/3 & 3/3/3 & 3/3/3 & 4/3/3\\
      Dynamic M & 4/5/4 & 5/5/5 & 5/5/4 & 7/7/6\\
      Dynamic NK & 7/6/- & 7/7/- & 7/7/- & 6/6/-\\
      \bottomrule
    \end{tabular}
  }
  \label{tbl:stablility_analysis_model}
\end{table}

\paragraph{Customized Kernel Latency.}
We further measure the throughput of the custom kernels for dynamic \textbf{M}, dynamic \textbf{NK}, and static \textbf{K} under different inputs, reporting TFLOPS in Tables~\ref{tbl:mlp_tflops_staircase_m}, \ref{tbl:attn_prefill_tflops_by_seqlen}, and~\ref{tbl:attn_decode_tflops_by_batch_size}. For GEMM, static \textbf{K} generally needs \textbf{M}=128 to match the dense baseline, although it surpasses dense at \textbf{M}=8 when \textbf{N} is large. Dynamic \textbf{M} with online skipping starts to outperform dense only when \textbf{M} is above 512. Once \textbf{M} exceeds 8192, the two sparse GEMM variants reach peak speedups of 1.35x and 2.20x, respectively. Attention shows a similar pattern in prefill: both sparse operators need about 32k query tokens to reach their best speedups, while sparse attention's block-wise \textbf{QK}$^\top$ preprocessing results in a 21\% gap relative to dynamic \textbf{M}. In decoding, dynamic \textbf{M} remains close to a 2x speedup under all settings because the online skipping can be processed token by token, similar to the situation in GEMV.

\begin{figure*}[!t]
  \centering
  \includegraphics[width=1.0\linewidth]{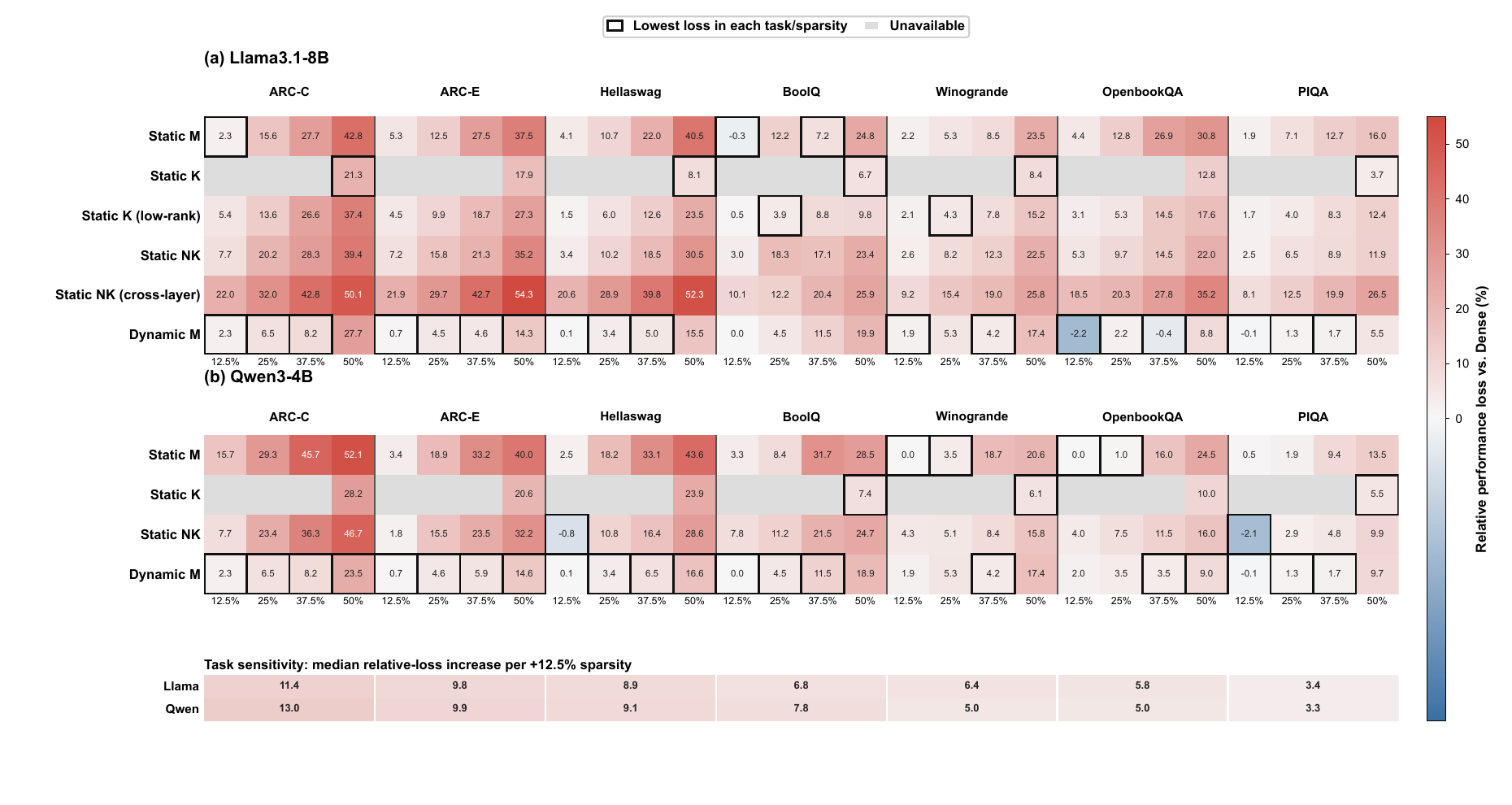}
  \caption{Relative performance loss across tasks, sparsity levels, and pruning taxonomies on Llama3.1-8B and Qwen3-4B. Tasks are ordered by their average sensitivity to sparsity.}
  \label{fig:task_wise_analysis}
\end{figure*}

\paragraph{Task-wise Sensitive Analysis.}
We further examine task-specific sensitivity to sparsity, as shown in Figure~\ref{fig:task_wise_analysis}. ARC-C, ARC-E, and HellaSwag are the most sensitive tasks, with their average relative performance loss approaching or exceeding 10\% for each 12.5\% increase in sparsity. By contrast, PIQA is less sensitive, with several pruning taxonomies retaining over 90\% performance at four sparsity levels. Across taxonomies, dynamic \textbf{M} performs best in most settings when excluding static \textbf{K}, whose results are concentrated at 50\% sparsity, where static \textbf{M} can remain competitive at low sparsity for tasks like ARC-C, Winogrande, and OpenbookQA.

\begin{figure*}[!ht]
  \centering
  \makebox[\linewidth][l]{
    \includegraphics[width=1.0\linewidth,trim=10 0 10 20,clip]{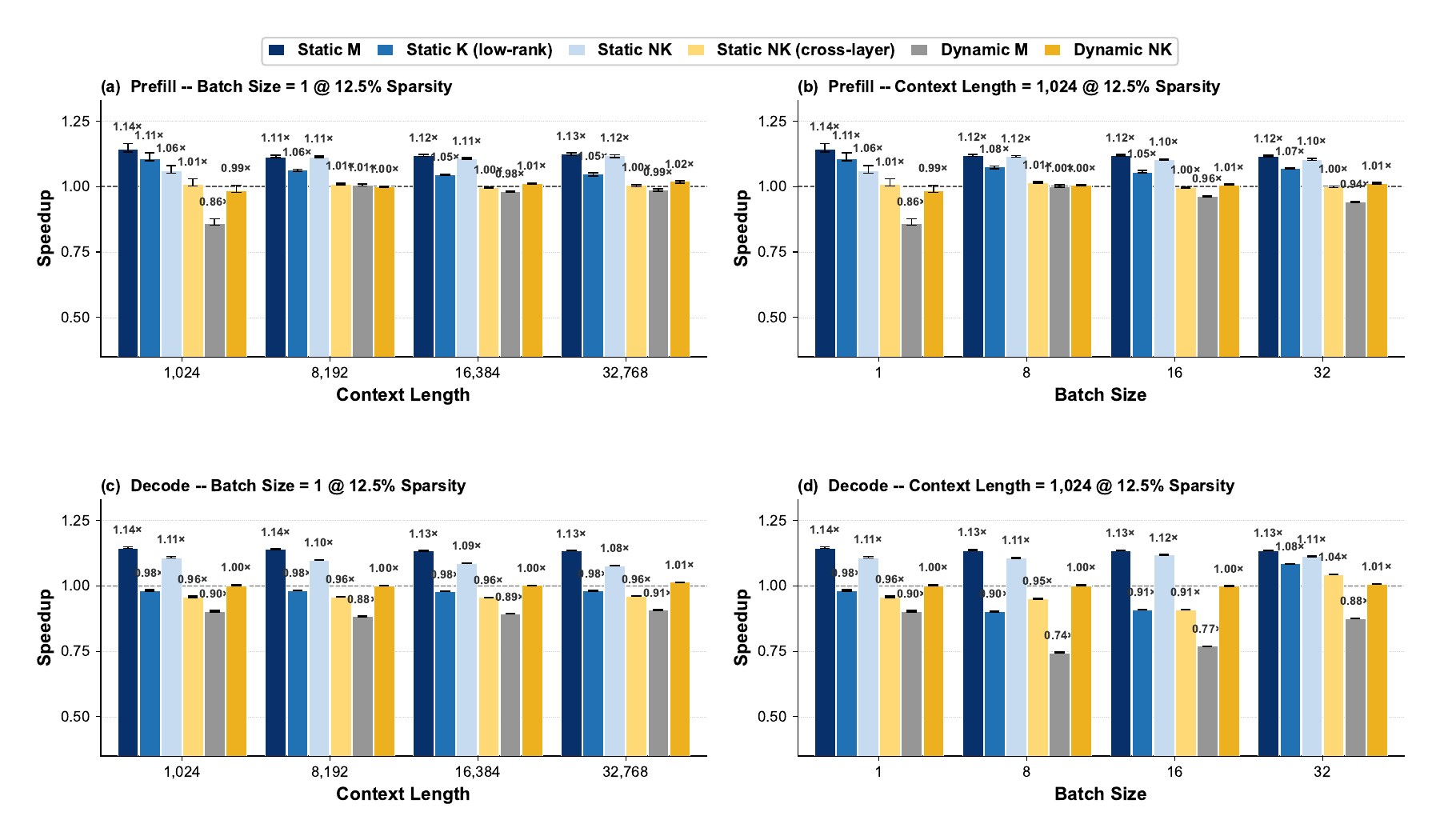}
  }
  \caption{Inference speedup on Llama3.1-8B with different pruning strategies, with sparsity set to 12.5\%.}
  \label{fig:ttft_tpot_0125_sm120}
\end{figure*}

\begin{figure*}[!ht]
  \centering
  \makebox[\linewidth][l]{
    \includegraphics[width=1.0\linewidth,trim=10 0 10 20,clip]{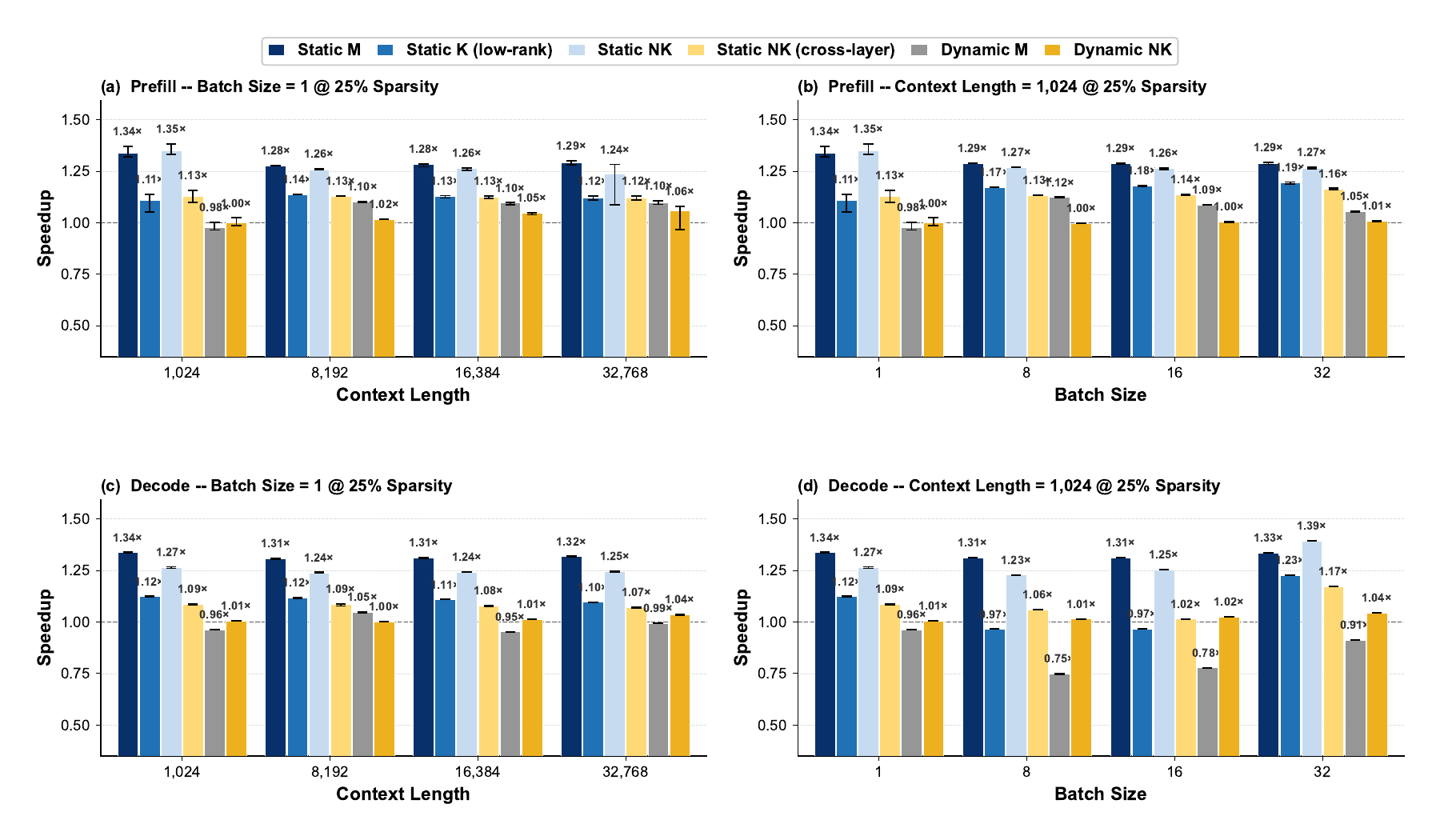}
  }
  \caption{Inference speedup on Llama3.1-8B with different pruning strategies, with sparsity set to 25\%.}
  \label{fig:ttft_tpot_025_sm120}
\end{figure*}

\begin{figure*}[!ht]
  \centering
  \makebox[\linewidth][l]{
    \includegraphics[width=1.0\linewidth,trim=10 0 10 20,clip]{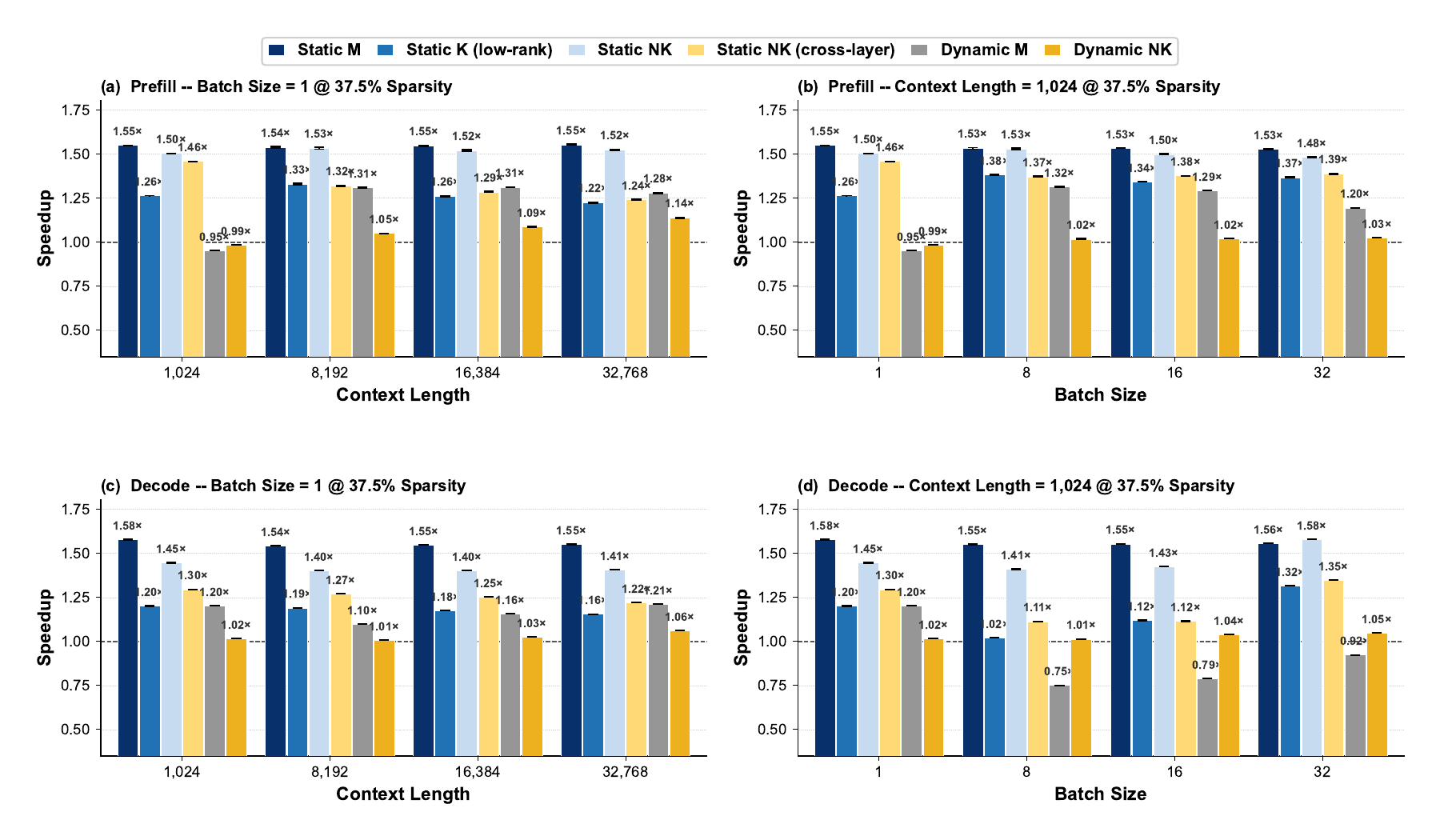}
  }
  \caption{Inference speedup on Llama3.1-8B with different pruning strategies, with sparsity set to 37.5\%.}
  \label{fig:ttft_tpot_0375_sm120}
\end{figure*}

\begin{figure*}[!ht]
  \centering
  \makebox[\linewidth][l]{
    \includegraphics[width=1.0\linewidth,trim=10 0 10 0,clip]{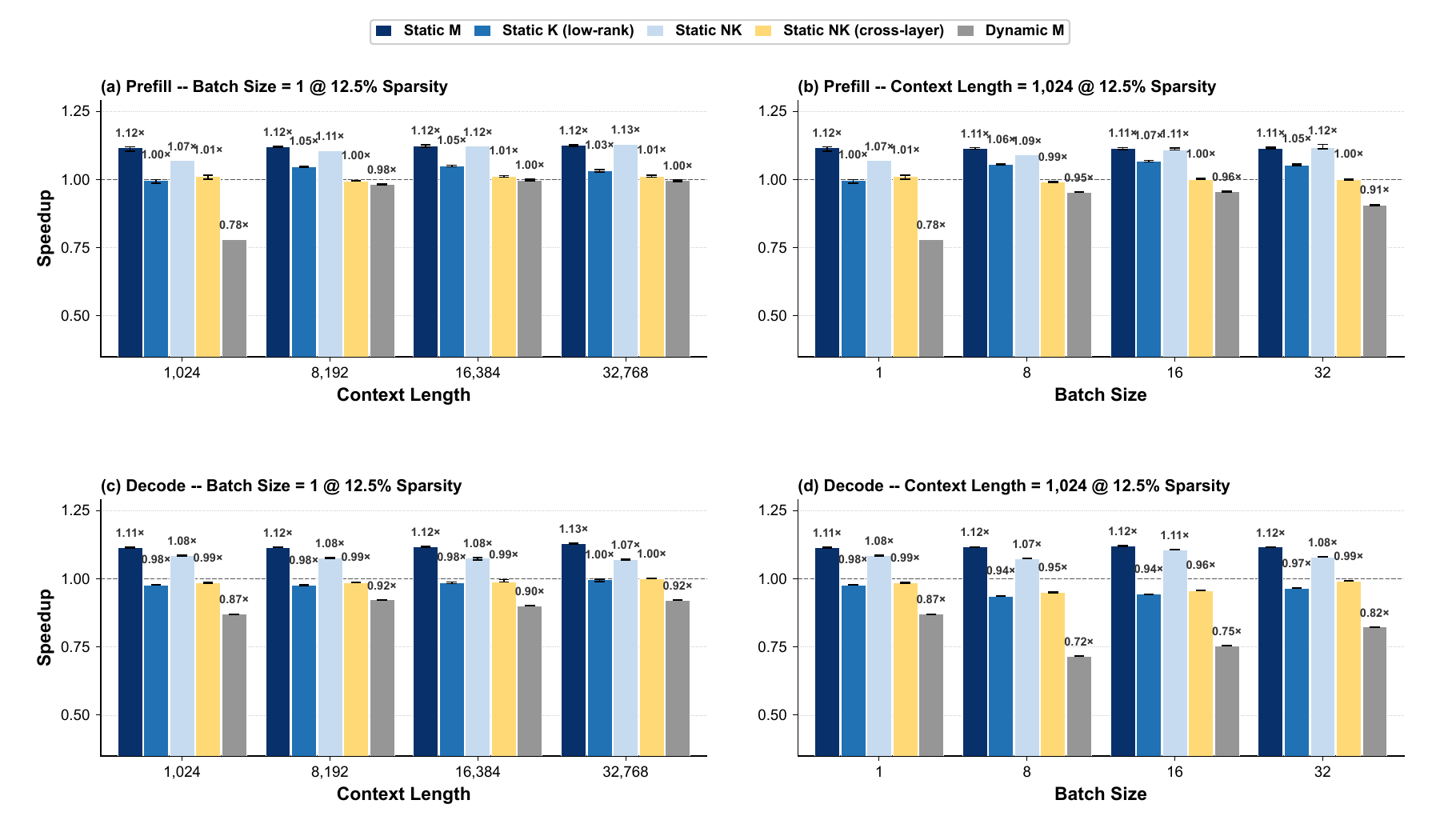}
  }
  \caption{Inference speedup on Qwen3-4B with different pruning strategies, with sparsity set to 12.5\%.}
  \label{fig:ttft_tpot_Qwen3_0125_sm120}
\end{figure*}

\begin{figure*}[!ht]
  \centering
  \makebox[\linewidth][l]{
    \includegraphics[width=1.0\linewidth,trim=10 0 10 0,clip]{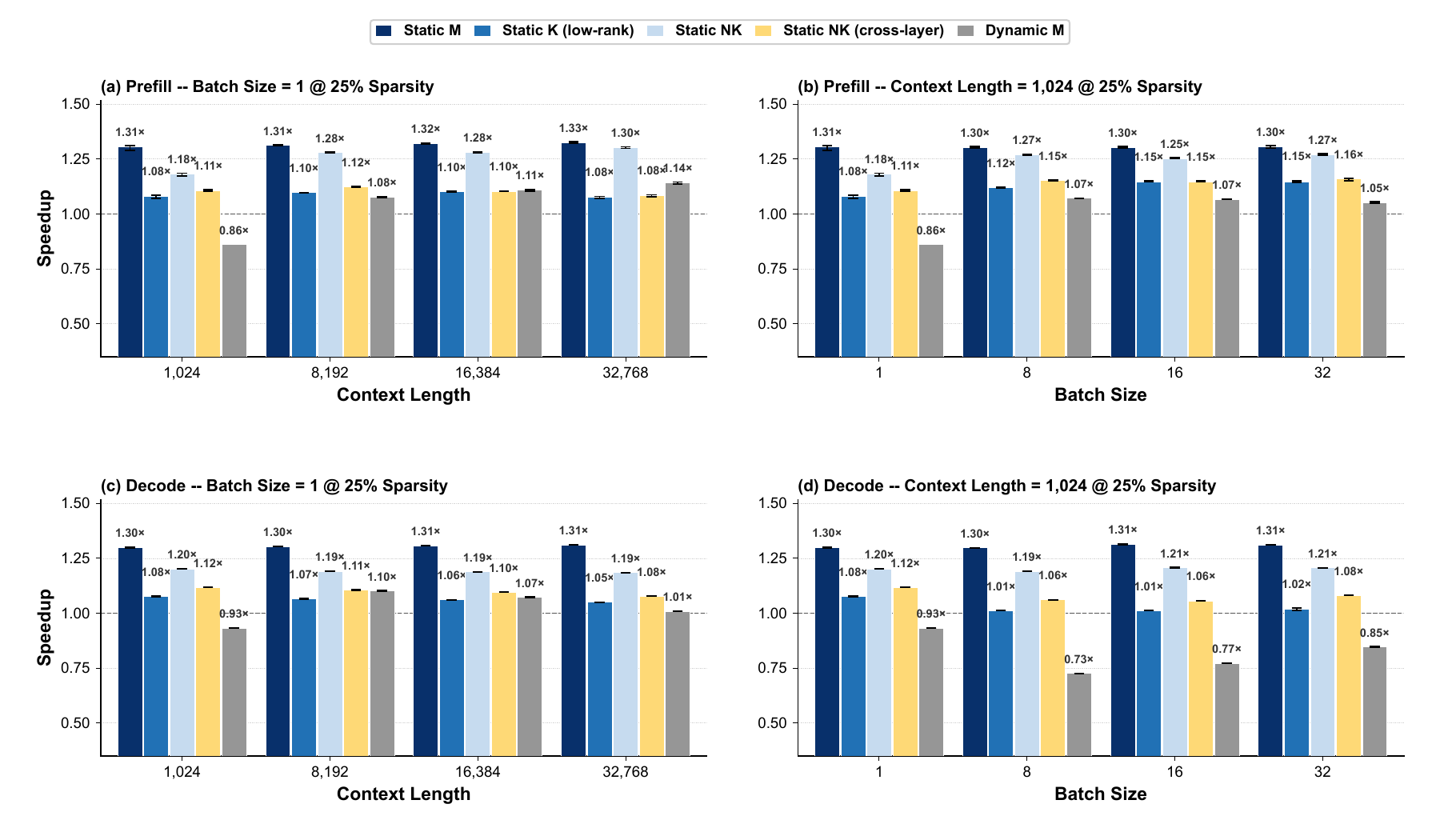}
  }
  \caption{Inference speedup on Qwen3-4B with different pruning strategies, with sparsity set to 25\%.}
  \label{fig:ttft_tpot_Qwen3_025_sm120}
\end{figure*}

\begin{figure*}[!ht]
  \centering
  \makebox[\linewidth][l]{
    \includegraphics[width=1.0\linewidth,trim=10 0 10 0,clip]{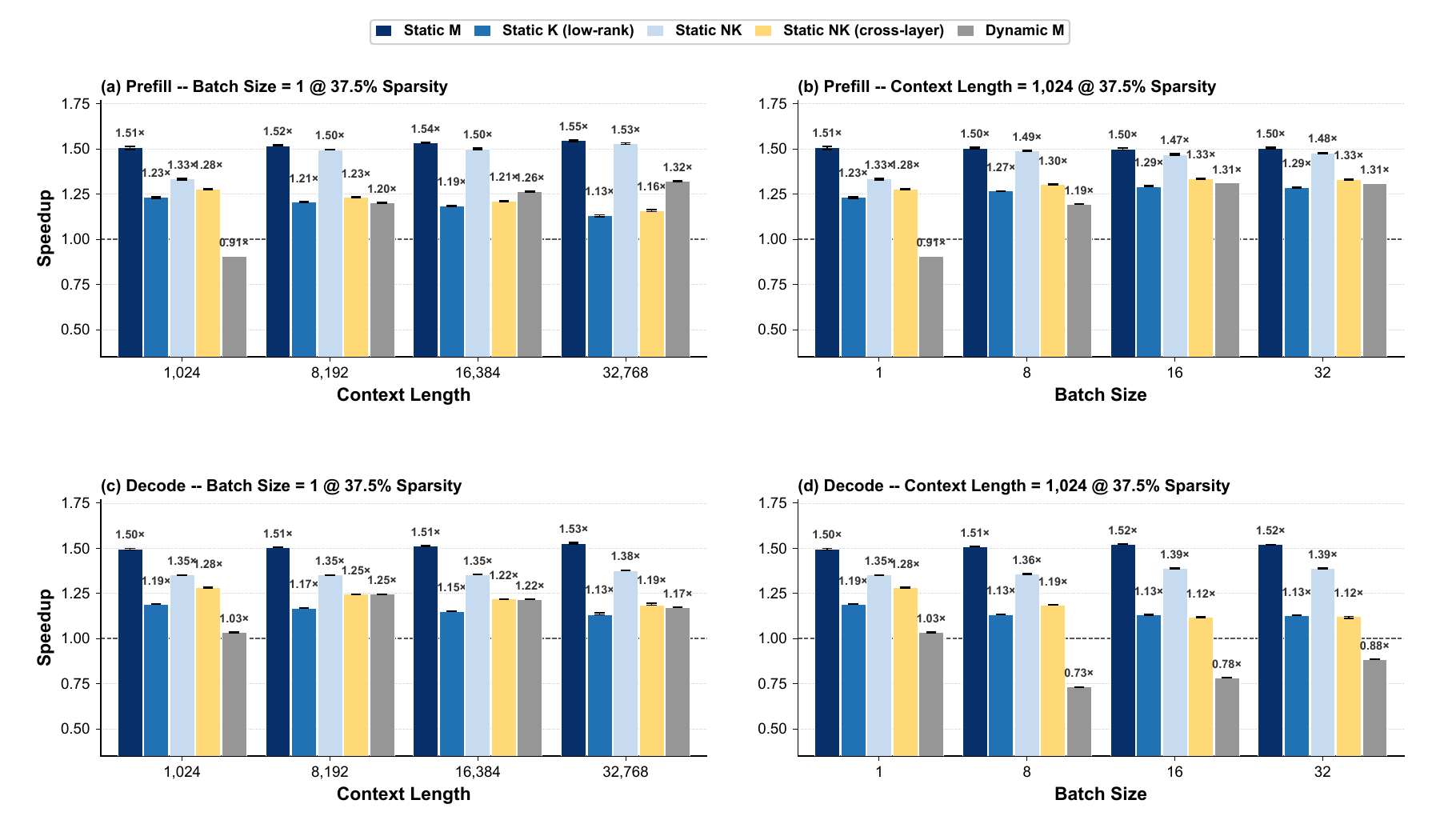}
  }
  \caption{Inference speedup on Qwen3-4B with different pruning strategies, with sparsity set to 37.5\%.}
  \label{fig:ttft_tpot_Qwen3_0375_sm120}
\end{figure*}

\begin{figure*}[!ht]
  \centering
  \makebox[\linewidth][l]{
    \includegraphics[width=1.0\linewidth,trim=10 0 10 0,clip]{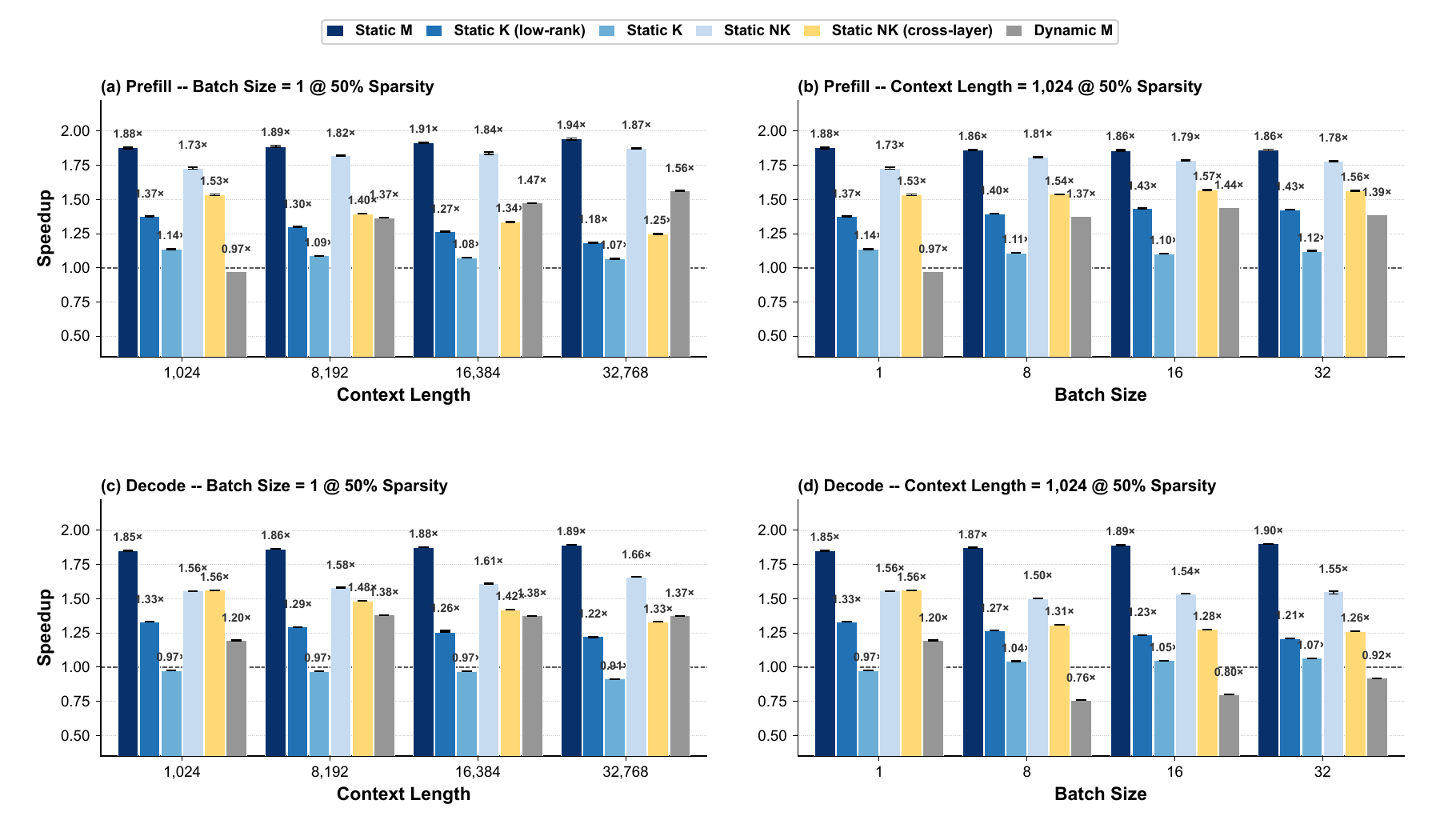}
  }
  \caption{Inference speedup on Qwen3-4B with different pruning strategies, with sparsity set to 50\%.}
  \label{fig:ttft_tpot_Qwen3_050_sm120}
\end{figure*}

\begin{figure*}[htbp]
  \centering
  \makebox[\linewidth][l]{
    \includegraphics[width=1.0\linewidth,trim=10 0 10 20,clip]{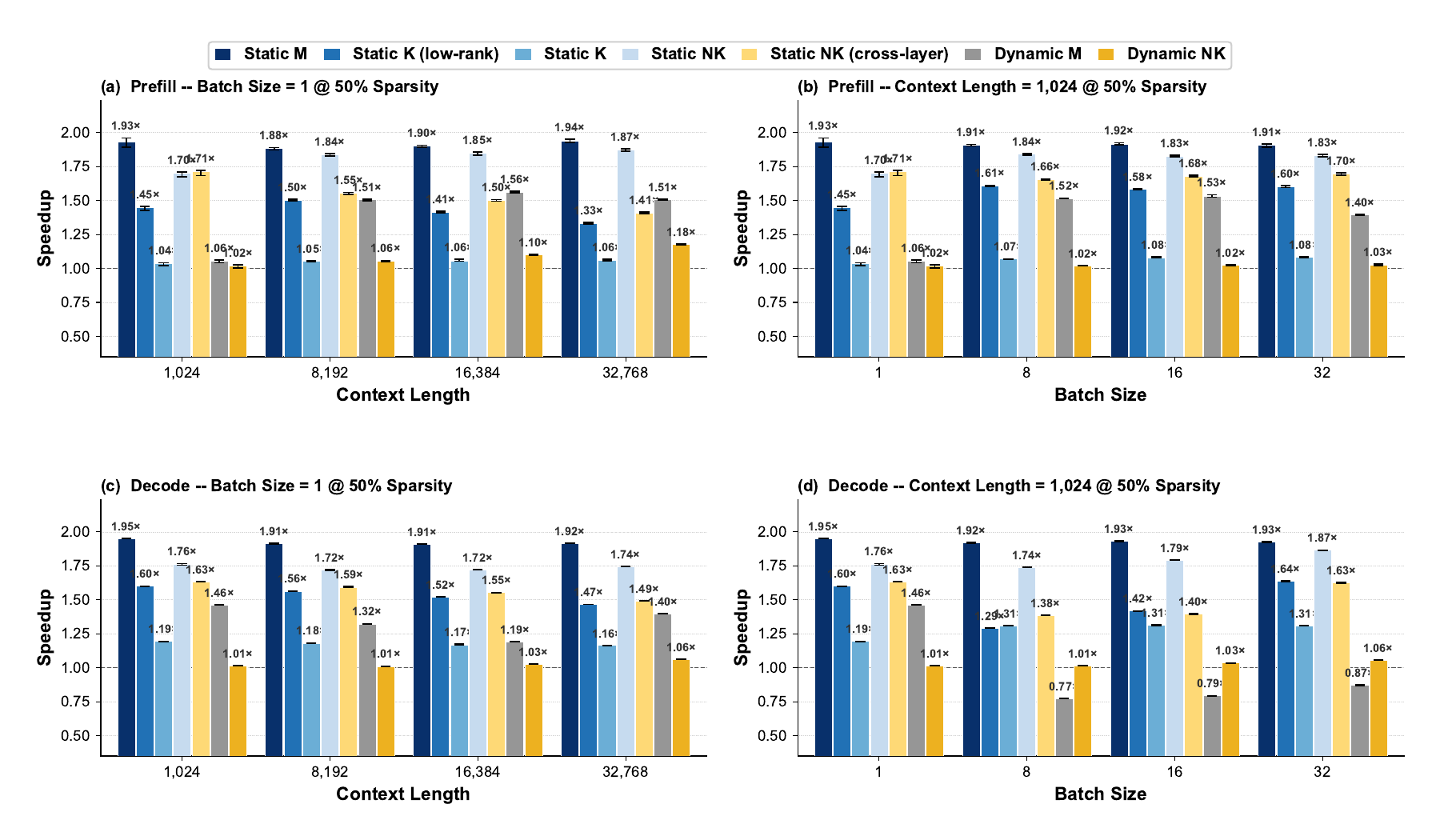}
  }
  \caption{Inference speedup on Qwen3-14B with different pruning strategies, with sparsity set to 50\%.}
  \label{fig:ttft_tpot_qwen3_14b_sm120}
\end{figure*}

\begin{figure*}[!ht]
  \centering
  \includegraphics[width=1.0\linewidth,trim=10 0 10 20,clip]{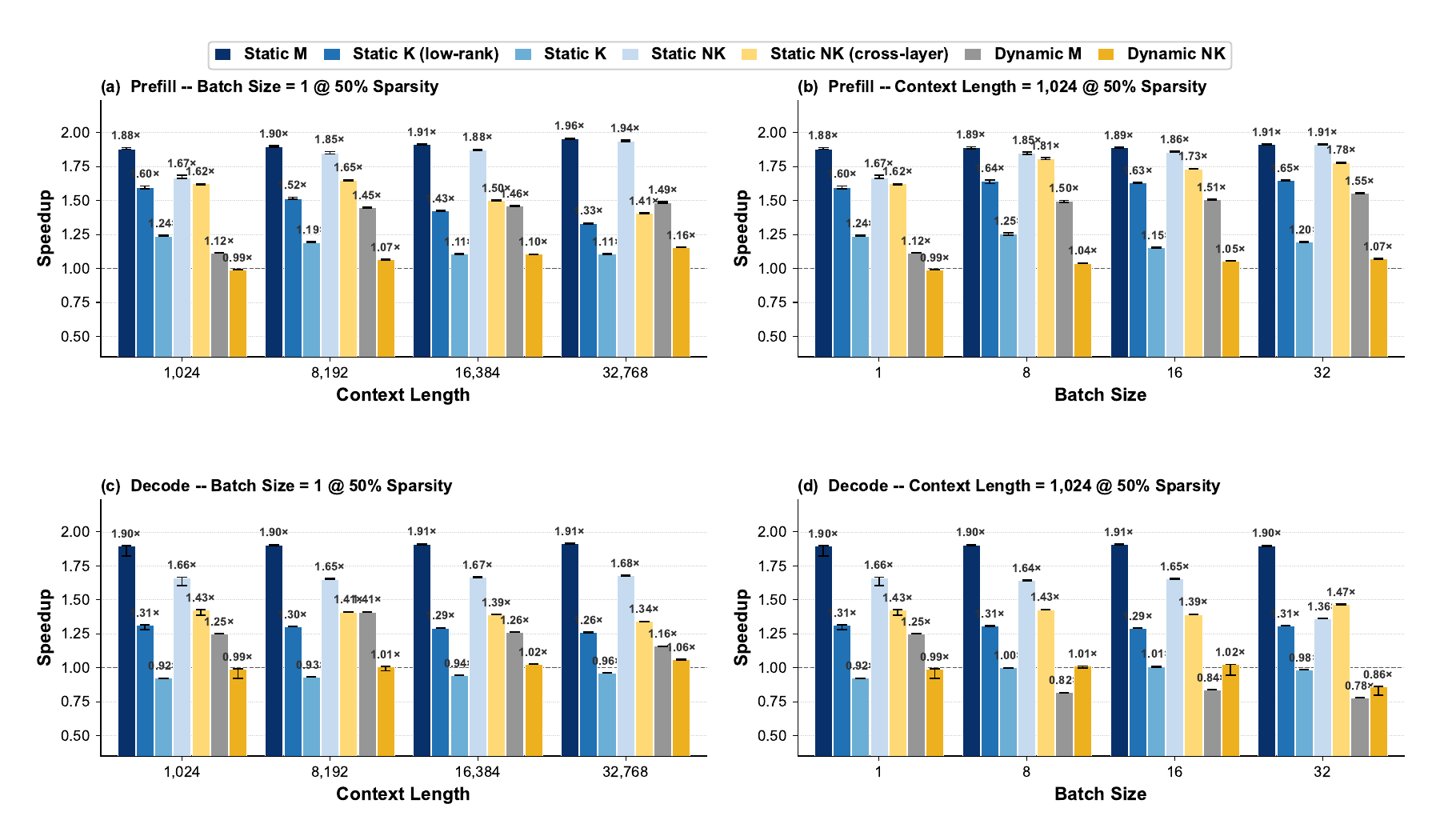}
  \caption{Inference speedup on Llama3.1-8B with different pruning strategies, with sparsity set to A800-80G.}
  \label{fig:ttft_tpot_sm80}
\end{figure*}

\begin{figure*}[!ht]
  \centering
  \includegraphics[width=1.0\linewidth,trim=10 0 10 0,clip]{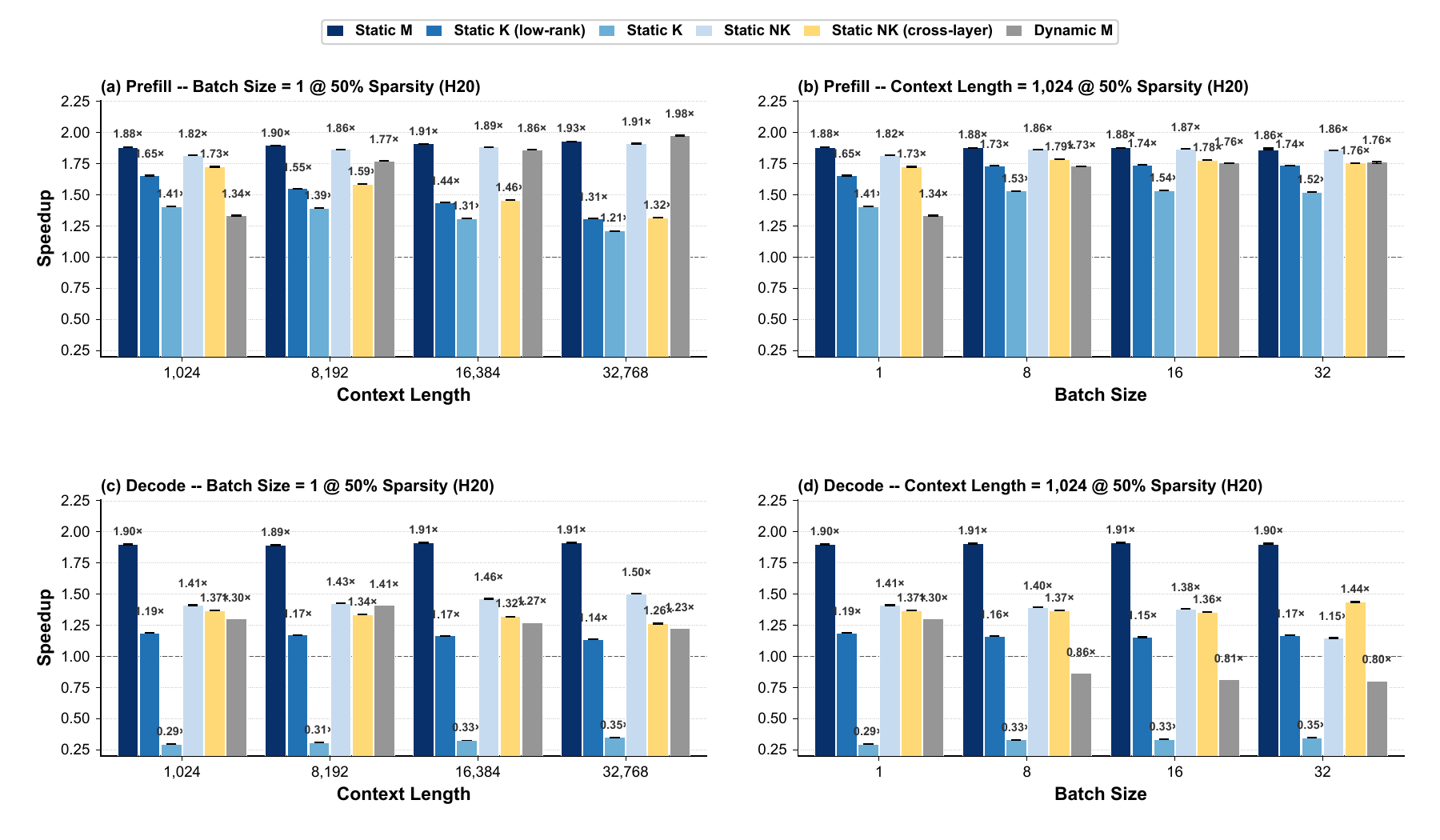}
  \caption{Inference speedup on Llama3.1-8B with different pruning strategies, with sparsity set to H20-96G.}
  \label{fig:ttft_tpot_sm90}
\end{figure*}

\paragraph{Experiments on Different Sparsity, Models, and Hardware.}
We also evaluate Llama3.1-8B on RTX Pro6000 at 12.5\%/25\%/37.5\% sparsity (Figure~\ref{fig:ttft_tpot_0125_sm120} to ~\ref{fig:ttft_tpot_0375_sm120}), Qwen3-4B on RTX Pro6000 at 12.5\%/25\%/37.5\%/50\% sparsity (Figure~\ref{fig:ttft_tpot_Qwen3_0125_sm120} to~\ref{fig:ttft_tpot_Qwen3_050_sm120}), Qwen3-14B at 50\% sparsity (Figure~\ref{fig:ttft_tpot_qwen3_14b_sm120}), and Llama3.1-8B on A800-80G (Figure~\ref{fig:ttft_tpot_sm80}) and H20-96G (Figure~\ref{fig:ttft_tpot_sm90}). These results follow the same overall trend as the experiments in main text (Table~\ref{tbl:stablility_analysis_hardware} and Table~\ref{tbl:stablility_analysis_model}).

\begin{table*}[!t]
  \centering
  \resizebox{\linewidth}{!}{
    \begin{tabular}{l c cc cc cc}
      \toprule
      \multirow{2}{*}{\textbf{Range}} & \multirow{2}{*}{\textbf{Align}} & \multicolumn{2}{c}{\textbf{Static K (low-rank)}} & \multicolumn{2}{c}{\textbf{Static NK}} & \multicolumn{2}{c}{\textbf{Static NK (cross-layer)}}\\
      \cmidrule(lr){3-4} \cmidrule(lr){5-6} \cmidrule(lr){7-8}
      & & Speed & Param. Ratio & Speed & Param. Ratio & Speed & Param. Ratio\\
      \midrule
      $\pm$ 0\% & -- & 19259.84 $\pm$ 28 & 56.73\% & 28334.09 $\pm$ 41 & 56.55\% & 20892.26 $\pm$ 13 & 53.35\%\\
      $\pm$ 20\% & odd & 0.7034 $\pm$ 0.0038 & 56.11\% & 0.7069 $\pm$ 0.0323 & 55.42\% & 0.6552 $\pm$ 0.0088 & 50.72\%\\
      $\pm$ 20\% & even & 0.9019 $\pm$ 0.0061 & 56.14\% & 0.9614 $\pm$ 0.0421 & 54.67\% & 0.8533 $\pm$ 0.0020 & 53.94\%\\
      $\pm$ 20\% & 4 & 0.9279 $\pm$ 0.0057 & 56.14\% & 0.9961 $\pm$ 0.0238 & 54.84\% & 0.8762 $\pm$ 0.0057 & 53.41\%\\
      $\pm$ 20\% & 8 & 0.9878 $\pm$ 0.0090 & 56.24\% & 1.0572 $\pm$ 0.0240 & 55.21\% & 0.9664 $\pm$ 0.0250 & 53.14\%\\
      $\pm$ 20\% & 16 & 1.0048 $\pm$ 0.0080 & 56.37\% & 1.0669 $\pm$ 0.0264 & 55.79\% & 0.9794 $\pm$ 0.0243 & 53.42\%\\
      $\pm$ 20\% & 128 & 0.9946 $\pm$ 0.0117 & 58.21\% & 1.0654 $\pm$ 0.0277 & 55.83\% & 0.9852 $\pm$ 0.0106 & 54.49\%\\
      \bottomrule
    \end{tabular}
  }
  \caption{Prefill speed and retained parameters under different pruning and alignment strategies at batch size 1, context length 32,768, and 50\% sparsity, where the 0\% row reports absolute throughput in token/s and the remaining rows report speed normalized by the corresponding 0\% baseline. Here, a 20\% range denotes the pruning probability of each unit randomly fluctuates by 20\% around the target sparsity, and align = 4 denotes alignment pruning dimensions to the nearest multiple of 4.}
  \label{tbl:static_width_alignment}
\end{table*}

\begin{table*}[htbp]
  \centering
  \resizebox{\linewidth}{!}{
    \begin{tabular}{l cccccccc}
      \toprule
      \multirow{2}{*}{\textbf{Pruning Dimension}} & \multicolumn{8}{c}{\textbf{Align}} \\
      \cmidrule(lr){2-9}
      & -- & odd & even & 4 & 8 & 16 & 32 & 64 \\
      \midrule
      \multicolumn{9}{c}{\textbf{bf16}} \\
      \midrule
      \textbf{N} & $345.26 \pm 0.49$ & $0.9280 \pm 0.0028$ & $0.9177 \pm 0.0017$ & $0.9758 \pm 0.0012$ & $0.9761 \pm 0.0011$ & $0.9867 \pm 0.0011$ & $0.9857 \pm 0.0015$ & $0.9910 \pm 0.0021$ \\
      \textbf{K} & $345.26 \pm 0.49$ & $0.3316 \pm 0.0023$ & $0.3342 \pm 0.0007$ & $0.3359 \pm 0.0021$ & $0.3371 \pm 0.0027$ & $1.0386 \pm 0.0219$ & $1.0112 \pm 0.0026$ & $1.0063 \pm 0.0023$ \\
      \midrule
      \multicolumn{9}{c}{\textbf{fp8\_e5m2}} \\
      \midrule
      \textbf{N} & $491.92 \pm 1.35$ & $0.9923 \pm 0.0029$ & $0.9925 \pm 0.0034$ & $0.9921 \pm 0.0027$ & $0.9932 \pm 0.0025$ & $0.9999 \pm 0.0031$ & $0.9978 \pm 0.0029$ & $0.9936 \pm 0.0027$ \\
      \textbf{K} & $491.92 \pm 1.35$ & $0.1181 \pm 0.0003$ & $0.1176 \pm 0.0003$ & $0.1182 \pm 0.0003$ & $0.1177 \pm 0.0003$ & $0.7043 \pm 0.0018$ & $0.9742 \pm 0.0026$ & $1.0460 \pm 0.0026$ \\
      \bottomrule
    \end{tabular}
  }
  \caption{Throughput of Triton's GEMM kernel under different alignment strategies, the base layout is set to [8192,8192]@[8192,8192], with \textbf{N}-axis align to the nearest odd denotes to apply [8192,8192]@[8192-1,8192] shape's GEMM.}
  \label{tbl:triton_gemm_align_test}
\end{table*}

\begin{table*}[htbp]
  \centering
  \begin{tabular}{l cc cc cc}
    \toprule
    \multirow{2}{*}{\textbf{Sparsity}} & \multicolumn{2}{c}{[4096,4096]@[4096,4096]} & \multicolumn{2}{c}{[4096,4096]@[14336,4096]} & \multicolumn{2}{c}{[4096,14336]@[4096,14336]}\\
    \cmidrule(lr){2-3} \cmidrule(lr){4-5} \cmidrule(lr){6-7}
    & \textbf{N} & \textbf{K} & \textbf{N} & \textbf{K} & \textbf{N} & \textbf{K}\\
    \midrule
    10\% & 1.0032 & 1.0963 & 1.1056 & 1.0900 & 1.0169 & 1.1009\\
    20\% & 1.0054 & 1.2189 & 1.2450 & 1.2265 & 1.0202 & 1.2406\\
    30\% & 1.0111 & 1.3613 & 1.4189 & 1.4001 & 1.4686 & 1.4306\\
    40\% & 1.4717 & 1.5739 & 1.6569 & 1.6232 & 1.5201 & 1.6674\\
    50\% & 1.4832 & 1.8469 & 1.9802 & 1.9223 & 1.9186 & 1.9752\\
    60\% & 1.4831 & 2.2075 & 2.4407 & 2.3059 & 2.2795 & 2.4348\\
    70\% & 2.4461 & 2.7465 & 3.2071 & 2.9513 & 2.9141 & 3.1823\\
    \bottomrule
  \end{tabular}
  \caption{The speedup provided by two different pruning dimensions on the query, up, and down projections of Llama3.1-8B under different sparsity. The shape of each GEMM is [\textbf{M}, \textbf{K}] @ [\textbf{N}, \textbf{K}].}
  \label{tbl:static_nk_vs_static_kn_in_mlp_prefill}
\end{table*}

\begin{table*}[htbp]
  \centering
  \begin{tabular}{l cc cc cc}
    \toprule
    \multirow{2}{*}{\textbf{Sparsity}} & \multicolumn{2}{c}{[1,4096]@[4096,4096]} & \multicolumn{2}{c}{[1,4096]@[14336,4096]} & \multicolumn{2}{c}{[1,14336]@[4096,14336]}\\
    \cmidrule(lr){2-3} \cmidrule(lr){4-5} \cmidrule(lr){6-7}
    & \textbf{N} & \textbf{K} & \textbf{N} & \textbf{K} & \textbf{N} & \textbf{K}\\
    \midrule
    10\% & 1.0119 & 0.9598 & 1.2674 & 0.9504 & 1.2619 & 1.1638\\
    20\% & 1.1205 & 1.0834 & 1.4695 & 1.1414 & 1.5728 & 1.5016\\
    30\% & 1.3939 & 1.1993 & 1.8296 & 1.3321 & 2.3106 & 1.9335\\
    40\% & 1.5471 & 1.2900 & 2.0408 & 1.6442 & 1.6629 & 2.0508\\
    50\% & 1.6532 & 1.7195 & 2.5662 & 2.3590 & 2.1222 & 2.9063\\
    60\% & 1.6396 & 1.6520 & 2.9301 & 2.6541 & 2.5912 & 3.2214\\
    70\% & 1.9354 & 1.8561 & 3.3894 & 3.1360 & 3.0271 & 3.0392\\
    \bottomrule
  \end{tabular}
  \caption{The speedup provided by two types of width pruning on the query, up, and down projections of Llama3.1-8B under different sparsity. The number of tokens is fixed to 1.}
  \label{tbl:static_nk_vs_static_kn_in_mlp_decode}
\end{table*}

\begin{table*}[htbp]
  \centering
  \resizebox{\linewidth}{!}{
    \begin{tabular}{l ccc ccc ccc}
      \toprule
      \multirow{2}{*}{\textbf{M}} & \multicolumn{3}{c}{[T,4096]@[4096,4096]} & \multicolumn{3}{c}{[T,4096]@[14336,4096]} & \multicolumn{3}{c}{[T,14336]@[4096,14336]}\\
      \cmidrule(lr){2-4} \cmidrule(lr){5-7} \cmidrule(lr){8-10}
      & dense & static \textbf{K} & dynamic \textbf{M} & dense & static \textbf{K} & dynamic \textbf{M} & dense & static \textbf{K} & dynamic \textbf{M}\\
      \midrule
      1     & 1.58   & 1.38   & 2.67 / 15.79 & 1.49   & 4.75   & 5.31 / 52.85 & 1.45   & 1.60   & 3.72 / 55.33 \\
      2     & 4.75   & 2.77   & 2.88          & 10.15  & 9.56   & 6.48          & 8.79   & 3.19   & 6.15          \\
      4     & 9.48   & 5.53   & 5.77          & 20.96  & 19.60  & 12.98         & 16.04  & 6.39   & 12.21         \\
      8     & 18.37  & 12.03  & 11.54         & 39.74  & 43.45  & 26.48         & 19.36  & 13.14  & 24.26         \\
      16    & 34.31  & 24.06  & 22.88         & 76.98  & 87.11  & 53.17         & 34.85  & 26.32  & 48.00         \\
      32    & 97.80  & 48.06  & 43.73         & 111.45 & 174.00 & 79.32         & 140.79 & 52.70  & 87.71         \\
      64    & 164.53 & 95.27  & 70.66         & 227.04 & 344.09 & 126.22        & 229.31 & 104.84 & 115.38        \\
      128   & 195.04 & 194.42 & 127.54        & 260.73 & 278.14 & 223.23        & 234.64 & 218.25 & 201.35        \\
      256   & 259.69 & 361.05 & 180.41        & 266.97 & 368.52 & 322.42        & 266.89 & 356.99 & 266.41        \\
      512   & 289.43 & 323.17 & 309.29        & 288.08 & 418.94 & 383.31        & 269.66 & 296.05 & 436.72        \\
      1024  & 292.02 & 407.81 & 231.15        & 325.65 & 427.26 & 513.34        & 278.75 & 394.01 & 261.52        \\
      2048  & 337.75 & 413.44 & 447.75        & 329.33 & 428.45 & 626.02        & 287.21 & 395.99 & 503.85        \\
      4096  & 303.37 & 432.04 & 599.81        & 300.49 & 431.03 & 640.65        & 289.59 & 412.07 & 640.92        \\
      8192  & 344.71 & 423.56 & 626.52        & 342.33 & 432.99 & 638.68        & 349.08 & 417.00 & 649.07        \\
      16384 & 324.97 & 426.68 & 652.30        & 292.01 & 433.49 & 649.14        & 292.42 & 420.96 & 644.12        \\
      32768 & 318.38 & 428.42 & 621.40        & 319.03 & 433.10 & 624.63        & 292.71 & 422.48 & 627.79        \\
      \bottomrule
    \end{tabular}
  }
  \caption{TFLOPS of dense, static \textbf{K}, and dynamic \textbf{M}'s GEMM kernels under different input sizes \textbf{M}, with sparsity fixed to 50\%. Static \textbf{K} reports the sglang JIT cuSPARSELt result, where dynamic M reports both no-skip and skip results when the batch size equals 1.}
  \label{tbl:mlp_tflops_staircase_m}
\end{table*}

\begin{table*}[htbp]
  \centering
  \begin{tabular}{l c c c}
    \toprule
    \multirow{2}{*}{\textbf{T}} & \multicolumn{3}{c}{[B,T,32,128]@[B,T,8,128]} \\
    \cmidrule(lr){2-4}
    & dense & dynamic \textbf{M} & dynamic \textbf{NK} \\
    \midrule
    1024  & 150.08 & 97.74  & 95.47  \\
    2048  & 207.57 & 197.44 & 186.35 \\
    4096  & 276.94 & 355.93 & 345.53 \\
    8192  & 320.49 & 483.24 & 433.20 \\
    16384 & 343.88 & 578.50 & 485.75 \\
    32768 & 351.15 & 634.31 & 523.63 \\
    65536 & 354.87 & 653.86 & 517.45 \\
    \bottomrule
  \end{tabular}
  \caption{Prefill TFLOPS of dense attention, dynamic \textbf{M}, and dynamic \textbf{NK} under different query lengths, with sparsity fixed to 50\%. The batch size is fixed to \textbf{1} where the shapes of the query, key, and value tensors are [batch size, context length, head counts, head size].}
  \label{tbl:attn_prefill_tflops_by_seqlen}
\end{table*}
\begin{table*}[htbp]
  \centering
  \begin{tabular}{l c c c}
    \toprule
    \multirow{2}{*}{\textbf{B}} & \multicolumn{3}{c}{[B,1,32,128]@[B,T,8,128]} \\
    \cmidrule(lr){2-4}
    & dense & dynamic \textbf{M} & dynamic \textbf{NK} \\
    \midrule
    1  & 7.89 & 8.16 / 52.92 & 8.47 \\
    2  & 2.55 & 10.61 & 7.58 \\
    4  & 2.63 & 5.75  & 4.93 \\
    8  & 2.66 & 5.94  & 5.26 \\
    16 & 2.69 & 6.04  & 5.00 \\
    32 & 3.05 & 6.09  & 6.00 \\
    \bottomrule
  \end{tabular}
  \caption{Decode TFLOPS of dense attention, dynamic \textbf{M}, and dynamic \textbf{NK} under different batch sizes, with sparsity fixed to 50\%. The KV cache length \textbf{T} is fixed to 32768, where dynamic \textbf{M} reports both no skip and skip results in batch size equals 1.}
  \label{tbl:attn_decode_tflops_by_batch_size}
\end{table*}

\begin{table*}[htbp]
    \centering
    \aboverulesep=0ex
    \belowrulesep=0ex
    \renewcommand{\arraystretch}{1.2}
    \resizebox{1.0\linewidth}{!}{
      \begin{tabular}{l|l |c|ccccccc|cc}
          \toprule
          \multirow{2}{*}{Categories} & \multirow{2}{*}{Methods} & WikiText2 & ARC-E & ARC-C & BoolQ & Winogrande & PIQA & OpenbookQA & Hellaswag & \multirow{2}{*}{Avg. Acc.$\uparrow$} & \multirow{2}{*}{Avg. Gap$\downarrow$}\\
          & & (ppl$\downarrow$) & (Acc. Norm.$\uparrow$) & (Acc. Norm.$\uparrow$) & (Acc. $\uparrow$) & (Acc. $\uparrow$) & (Acc. Norm.$\uparrow$) & (Acc. Norm.$\uparrow$) & (Acc. Norm.$\uparrow$) & &\\
          \midrule
          & Dense & 7.54 & 82.44 & 55.20 & 82.96 & 74.34 & 80.84 & 45.40 & 79.34 & 71.50 & 0.00\%\\
          & Dense (LoRA) & 7.74 & 81.35 & 54.18 & 83.48 & 72.69 & 81.22 & 46.60 & 79.94 & 71.35 & 0.21\%\\
          \midrule
          \multicolumn{12}{c}{\textbf{12.5\% Sparsity}}\\
          \midrule
          \multirow{2}{*}{Static \textbf{M}} & Shortened-taylor & 10.54 & 78.07 & \textbf{53.92} & \textbf{83.21} & 72.69 & 79.33 & 43.40 & 76.08 & 69.52 & 2.76\%\\
          & CoopPruner & 10.90 & 72.47 & 47.10 & 74.59 & 66.46 & 78.89 & 43.40 & 75.16 & 65.43 & 8.48\%\\
          \midrule
          \multirow{1}{*}{Static \textbf{K}} & Dobi-SVD$^\dagger$ & \underline{8.72} & \underline{78.75} & \underline{52.22} & 82.54 & \underline{72.77} & \underline{79.43} & \underline{44.00} & \underline{78.13} & \underline{69.69} & \underline{2.53\%}\\
          \midrule
          \multirow{2}{*}{Static \textbf{NK}} & Týr-the-Pruner & 8.81 & 76.52 & 50.94 & 80.46 & 72.38 & 78.78 & 43.00 & 76.62 & 68.38 & 4.35\%\\
          & SliceGPT$^\ddagger$ & 35.05 & 64.35 & 43.08 & 74.55 & 67.48 & 74.31 & 37.00 & 62.96 & 60.53 & 15.34\%\\
          \midrule
          \multirow{1}{*}{Dynamic \textbf{M}} & SkipGPT & \textbf{8.42} & \textbf{81.86} & \textbf{53.92} & \underline{82.96} & \textbf{72.92} & \textbf{80.90} & \textbf{46.40} & \textbf{79.26} & \textbf{71.17} & \textbf{0.45\%}\\
          \midrule
          \multicolumn{12}{c}{\textbf{25\% Sparsity}}\\
          \midrule
          \multirow{4}{*}{Static \textbf{M}} & Shortened-ppl & 13.61 & 68.73 & 42.75 & 64.77 & 63.54 & 76.99 & 42.40 & 70.38 & 61.36 & 14.17\%\\
          & Shortened-taylor & 15.52 & 72.10 & 46.59 & 72.81 & \underline{70.40} & 75.14 & 39.60 & 70.84 & 63.92 & 10.59\%\\
          & CoopPruner & 13.59 & 68.51 & 41.80 & 65.04 & 63.06 & 76.27 & 40.20 & 70.41 & 60.75 & 15.03\%\\
          & BlockPruner & 16.53 & 66.20 & 40.78 & 66.06 & 65.35 & 74.59 & 39.60 & 65.10 & 59.66 & 16.55\%\\
          \midrule
          \multirow{1}{*}{Static \textbf{K}} & Dobi-SVD$^\dagger$ & 10.14 & 74.24 & 47.70 & \textbf{79.69} & \textbf{71.11} & 77.58 & 43.00 & 74.56 & \underline{66.84} & \underline{6.52\%}\\
          \midrule
          \multirow{3}{*}{Static \textbf{NK}} & FLAP & 22.69 & 52.31 & 32.08 & 62.45 & 53.99 & 69.80 & 33.40 & 53.57 & 51.08 & 28.55\%\\
          & Týr-the-Pruner & 12.46 & 69.44 & 44.03 & 67.74 & 68.27 & 75.57 & 41.00 & 71.21 & 62.46 & 12.63\%\\
          & SliceGPT$^\ddagger$ & 38.96 & 57.95 & 37.54 & 72.87 & 62.90 & 70.73 & 36.20 & 56.40 & 56.37 & 21.16\%\\
          \midrule
          \multirow{2}{*}{Dynamic \textbf{M}} & D-LLM & 9.58 & \underline{76.85} & \underline{48.75} & 75.44 & 69.21 & \underline{77.85} & \underline{43.20} & \underline{74.67} & 66.56 & 6.90\%\\
          & SkipGPT & \underline{9.25} & \textbf{78.70} & \textbf{51.62} & \underline{79.26} & \underline{70.40} & \textbf{79.76} & \textbf{44.40} & \textbf{76.62} & \textbf{68.68} & \textbf{3.94\%}\\
          \midrule
          \multirow{1}{*}{Dynamic \textbf{NK}} & SeerAttention$^\dagger$ & \textbf{7.72} & 49.62 & 34.81 & 53.67 & 54.46 & 60.94 & 38.40 & 41.41 & 47.61 & 33.40\%\\
          \midrule
          \multicolumn{12}{c}{\textbf{37.5\% Sparsity}}\\
          \midrule
          \multirow{2}{*}{Static \textbf{M}} & Shortened-taylor & 26.52 & 59.76 & 39.93 & \textbf{76.97} & 68.03 & 70.57 & 33.20 & 61.88 & 58.62 & 18.01\%\\
          & CoopPruner & 23.03 & 59.43 & 34.22 & 60.55 & 60.14 & 72.03 & 34.80 & 58.21 & 54.19 & 24.20\%\\
          \midrule
          \multirow{1}{*}{Static \textbf{K}} & Dobi-SVD$^\dagger$ & \underline{12.34} & \underline{67.05} & \underline{40.53} & \underline{75.66} & \underline{68.51} & \underline{74.16} & \underline{38.80} & \underline{69.37} & \underline{62.01} & \underline{13.27\%}\\
          \midrule
          \multirow{2}{*}{Static \textbf{NK}} & Týr-the-Pruner & 26.18 & 64.90 & 39.59 & 68.78 & 65.19 & 73.61 & \underline{38.80} & 64.68 & 59.36 & 16.97\%\\
          & SliceGPT$^\ddagger$ & 57.51 & 47.26 & 31.56 & 66.05 & 60.22 & 64.79 & 32.80 & 47.77 & 50.06 & 29.98\%\\
          \midrule
          \multirow{1}{*}{Dynamic \textbf{M}} & SkipGPT & \textbf{10.64} & \textbf{78.66} & \textbf{50.68} & 73.39 & \textbf{71.19} & \textbf{79.43} & \textbf{45.60} & \textbf{75.37} & \textbf{67.76} & \textbf{5.23\%}\\
          \midrule
          \multicolumn{12}{c}{\textbf{50\% Sparsity}}\\
          \midrule
          \multirow{4}{*}{Static \textbf{M}} & Shortened-ppl & 36.43 & 48.74 & 28.75 & 61.50 & 52.57 & 65.29 & 29.60 & 44.25 & 47.24 & 33.92\%\\
          & Shortened-taylor & 52.92 & 47.18 & 31.48 & 64.40 & 59.27 & 65.61 & 28.80 & 48.80 & 49.36 & 30.96\%\\
          & CoopPruner & 33.93 & 51.55 & 31.56 & 62.41 & 56.90 & 67.89 & 31.40 & 47.24 & 49.85 & 30.28\%\\
          & BlockPruner & 71.96 & 40.82 & 25.51 & 55.05 & 53.12 & 61.53 & 27.60 & 37.38 & 43.00 & 39.86\%\\
          \midrule
          \multirow{2}{*}{Static \textbf{K}} & Dobi-SVD$^\dagger$ & 15.44 & 59.93 & 34.56 & \underline{74.83} & \underline{63.06} & 70.78 & 37.40 & 60.66 & 57.31 & 19.83\%\\
          & MaskLLM & \underline{11.45} & \underline{67.72} & \textbf{43.43} & \textbf{77.43} & \textbf{68.11} & \textbf{77.86} & \underline{39.60} & \textbf{72.95} & \textbf{63.87} & \textbf{10.67\%}\\
          \midrule
          \multirow{3}{*}{Static \textbf{NK}} & FLAP & 88.93 & 38.34 & 24.82 & 58.07 & 51.85 & 59.73 & 26.40 & 33.55 & 41.82 & 41.50\%\\
          & Týr-the-Pruner & 19.80 & 53.45 & 33.45 & 63.58 & 57.62 & 71.22 & 35.40 & 55.13 & 52.83 & 26.10\%\\
          & SliceGPT$^\ddagger$ & 68.71 & 37.67 & 27.56 & 61.44 & 55.17 & 59.41 & 29.40 & 37.86 & 44.07 & 38.36\%\\
          \midrule
          \multirow{2}{*}{Dynamic \textbf{M}} & D-LLM & 16.83 & 62.58 & 34.30 & 62.44 & 57.69 & 71.87 & 36.80 & 59.84 & 55.07 & 22.97\%\\
          & SkipGPT & 13.14 & \textbf{70.62} & \underline{39.93} & 66.42 & 61.40 & \underline{76.38} & \textbf{41.40} & \underline{67.05} & \underline{60.45} & \underline{15.44\%}\\
          \midrule
          \multirow{1}{*}{Dynamic \textbf{NK}} & SeerAttention$^*$ & \textbf{7.76} & 49.62 & 34.81 & 48.62 & 54.46 & 60.77 & 38.40 & 41.41 & 46.87 & 34.45\%\\
          \bottomrule
      \end{tabular}
    }
    \caption{Llama3.1-8B pruning performance under the LoRA configuration, where the static semi-structured \textbf{K} methods are reported without LoRA fine-tuning due to the lack of unified LoRA implementation strategies. $*$ for dynamic \textbf{NK} denotes applying the attention gate checkpoint for Llama3.1-8B-Instruct to the base Llama3.1-8B due to high retraining overhead.}
    \label{tbl:performance_lora}
\end{table*}

\begin{table*}[htbp]
    \centering
    \aboverulesep=0ex
    \belowrulesep=0ex
    \renewcommand{\arraystretch}{1.2}
    \resizebox{1.0\linewidth}{!}{
      \begin{tabular}{l|l |c|ccccccc|cc}
          \toprule
          \multirow{2}{*}{Categories} & \multirow{2}{*}{Methods} & WikiText2 & ARC-E & ARC-C & BoolQ & Winogrande & PIQA & OpenbookQA & Hellaswag & \multirow{2}{*}{Avg. Acc.$\uparrow$} & \multirow{2}{*}{Avg. Gap$\downarrow$}\\
          & & (ppl$\downarrow$) & (Acc. Norm.$\uparrow$) & (Acc. Norm.$\uparrow$) & (Acc. $\uparrow$) & (Acc. $\uparrow$) & (Acc. Norm.$\uparrow$) & (Acc. Norm.$\uparrow$) & (Acc. Norm.$\uparrow$) & &\\
          \midrule
          & Dense & 17.31 & 78.41 & 54.10 & 85.17 & 65.98 & 74.86 & 40.00 & 68.37 & 66.69 & 0.00\%\\
          \midrule
          \multicolumn{12}{c}{\textbf{12.5\% Sparsity}}\\
          \midrule
          \multirow{4}{*}{Static \textbf{M}} & Shortened-ppl & 14.42 & 72.31 & 44.03 & 66.45 & 59.19 & 75.35 & \textbf{41.00} & 63.20 & 60.21 & 9.71\%\\
          & Shortened-taylor & 15.77 & 75.78 & 45.60 & 82.39 & \textbf{65.98} & 74.51 & 40.00 & 66.63 & 64.41 & 3.42\%\\
          & CoopPruner & 14.47 & 72.69 & 44.62 & 65.87 & 61.72 & 74.92 & 40.00 & 63.25 & 60.43 & 9.38\%\\
          & ReplaceMe & 19.56 & 67.80 & 43.60 & 78.50 & \underline{65.75} & 70.73 & 37.40 & 59.80 & 60.51 & 9.27\%\\
          \midrule
          \multirow{2}{*}{Static \textbf{NK}} & LoRAP & 14.31 & 76.30 & 49.74 & 75.66 & 64.09 & \textbf{76.55} & 40.20 & 65.57 & 64.01 & 4.02\%\\
          & Týr-the-Pruner & \underline{12.72} & \underline{76.98} & 49.91 & 78.50 & 63.11 & \underline{76.44} & 38.40 & \textbf{68.91} & 64.60 & 3.13\%\\
          \midrule
          \multirow{2}{*}{Dynamic \textbf{M}} & MoD & 12.82 & 76.02 & \underline{52.59} & \underline{84.47} & 64.38 & 74.58 & \underline{40.80} & 68.04 & \underline{65.84} & \underline{1.28\%}\\
          & SkipGPT & \textbf{10.13} & \textbf{77.85} & \textbf{52.84} & \textbf{85.17} & 64.71 & 74.91 & 39.20 & \underline{68.30} & \textbf{66.14} & \textbf{0.83\%}\\
          \midrule
          \multicolumn{12}{c}{\textbf{25\% Sparsity}}\\
          \midrule
          \multirow{4}{*}{Static \textbf{M}} & Shortened-ppl & 16.94 & 61.20 & 36.43 & 59.42 & 55.49 & 73.39 & \underline{39.60} & 55.79 & 54.47 & 18.32\%\\
          & Shortened-taylor & 17.05 & 63.59 & 38.26 & 78.01 & \textbf{63.69} & 73.46 & \underline{39.60} & 55.93 & 58.93 & 11.64\%\\
          & CoopPruner & 17.29 & 63.13 & 37.54 & 51.01 & 54.93 & 73.12 & \textbf{40.40} & 55.25 & 53.62 & 19.59\%\\
          & ReplaceMe & 30.82 & 57.58 & 35.32 & 76.57 & 62.43 & 66.59 & 31.20 & 51.69 & 54.48 & 18.31\%\\
          \midrule
          \multirow{2}{*}{Static \textbf{NK}} & LoRAP & 18.03 & 65.64 & 40.62 & 73.91 & 61.25 & 73.23 & 34.20 & 58.23 & 58.15 & 12.81\%\\
          & Týr-the-Pruner & 14.83 & 66.25 & 41.45 & 75.63 & \underline{62.61} & 72.67 & 37.00 & 60.96 & 59.51 & 10.77\%\\
          \midrule
          \multirow{2}{*}{Dynamic \textbf{M}} & MoD & \underline{13.68} & \underline{73.70} & \underline{48.50} & \underline{80.71} & \underline{62.61} & \underline{73.61} & 38.20 & \underline{64.50} & \underline{63.11} & \underline{5.36\%}\\
          & SkipGPT & \textbf{12.52} & \textbf{74.83} & \textbf{50.59} & \textbf{81.37} & 62.49 & \textbf{73.85} & 38.59 & \textbf{66.02} & \textbf{63.96} & \textbf{4.09\%}\\
          \midrule
          \multicolumn{12}{c}{\textbf{37.5\% Sparsity}}\\
          \midrule
          \multirow{4}{*}{Static \textbf{M}} & Shortened-ppl & 26.32 & 51.73 & 29.52 & 58.99 & 52.25 & 67.68 & 34.00 & 44.83 & 48.42 & 27.39\%\\
          & Shortened-taylor & 45.34 & 46.00 & 31.74 & 66.73 & 58.48 & 63.66 & 30.20 & 43.08 & 48.55 & 27.20\%\\
          & CoopPruner & 26.03 & 52.36 & 29.35 & 58.20 & 53.67 & 67.85 & 33.60 & 45.71 & 48.67 & 27.01\%\\
          & ReplaceMe & 53.73 & 44.28 & 27.13 & 57.09 & 57.06 & 62.08 & 30.20 & 39.42 & 45.32 & 32.04\%\\
          \midrule
          \multirow{2}{*}{Static \textbf{NK}} & LoRAP & 25.18 & 60.14 & 35.15 & 68.04 & 57.85 & 68.99 & 34.40 & 48.76 & 53.33 & 20.03\%\\
          & Týr-the-Pruner & \underline{17.50} & 60.02 & 34.47 & 66.85 & \underline{60.46} & 71.27 & 35.40 & 57.16 & 55.09 & 17.40\%\\
          \midrule
          \multirow{2}{*}{Dynamic \textbf{M}} & MoD & 18.75 & \underline{72.25} & \underline{46.18} & \underline{74.78} & 60.27 & \underline{72.93} & \underline{36.20} & \underline{61.57} & \underline{60.59} & \underline{9.14\%}\\
          & SkipGPT & \textbf{15.89} & \textbf{73.76} & \textbf{49.67} & \textbf{75.34} & \textbf{63.18} & \textbf{73.55} & \textbf{38.60} & \textbf{63.94} & \textbf{62.58} & \textbf{6.17\%}\\
          \midrule
          \multicolumn{12}{c}{\textbf{50\% Sparsity}}\\
          \midrule
          \multirow{4}{*}{Static \textbf{M}} & Shortened-ppl & 37.89 & 47.05 & 25.94 & 60.86 & 52.41 & 64.74 & 30.20 & 38.53 & 45.67 & 31.51\%\\
          & Shortened-taylor & 61.50 & 43.10 & 25.77 & 55.20 & 54.06 & 61.59 & 30.40 & 36.26 & 43.76 & 34.37\%\\
          & CoopPruner & 36.03 & 47.14 & 27.22 & 52.54 & 51.85 & 64.04 & 31.40 & 39.13 & 44.76 & 32.89\%\\
          & ReplaceMe & 66.89 & 40.40 & 25.85 & 45.63 & 49.72 & 58.71 & 28.80 & 34.17 & 40.46 & 39.32\%\\
          \midrule
          \multirow{1}{*}{Static \textbf{K}} & HyperPrune$^\dagger$ & 28.85 & 62.25 & \underline{38.82} & \textbf{78.87} & \textbf{61.96} & \textbf{70.78} & \underline{36.00} & 52.05 & \textbf{57.24} & \textbf{14.17\%}\\
          \midrule
          \multirow{2}{*}{Static \textbf{NK}} & LoRAP & 36.40 & 51.14 & 29.61 & 64.46 & 53.51 & 62.51 & 32.00 & 38.95 & 47.45 & 28.85\%\\
          & Týr-the-Pruner & 21.44 & 53.20 & 28.84 & 64.10 & \underline{55.56} & 67.46 & 33.60 & 48.83 & 50.22 & 24.69\%\\
          \midrule
          \multirow{2}{*}{Dynamic \textbf{M}} & MoD & \underline{20.03} & \underline{64.01} & 32.79 & 66.50 & 54.93 & 63.20 & 32.60 & \textbf{57.51} & 53.07 & 20.42\%\\
          & SkipGPT & \textbf{19.17} & \textbf{66.98} & \textbf{41.37} & \underline{69.04} & 54.49 & \underline{67.59} & \textbf{36.40} & \underline{57.02} & \underline{56.13} & \underline{15.84\%}\\
          \bottomrule
      \end{tabular}
    }
    \caption{Qwen3-4B pruning performance under the LoRA configuration. Static \textbf{K} methods marked with $\dagger$ are reported without LoRA fine-tuning due to the lack of unified LoRA implementation strategies.}
    \label{tbl:performance_lora_qwen3_4b}
\end{table*}

\begin{figure*}[!ht]
  \centering
  \makebox[\linewidth][l]{
    \includegraphics[width=1.0\linewidth,trim=20 10 10 0,clip]{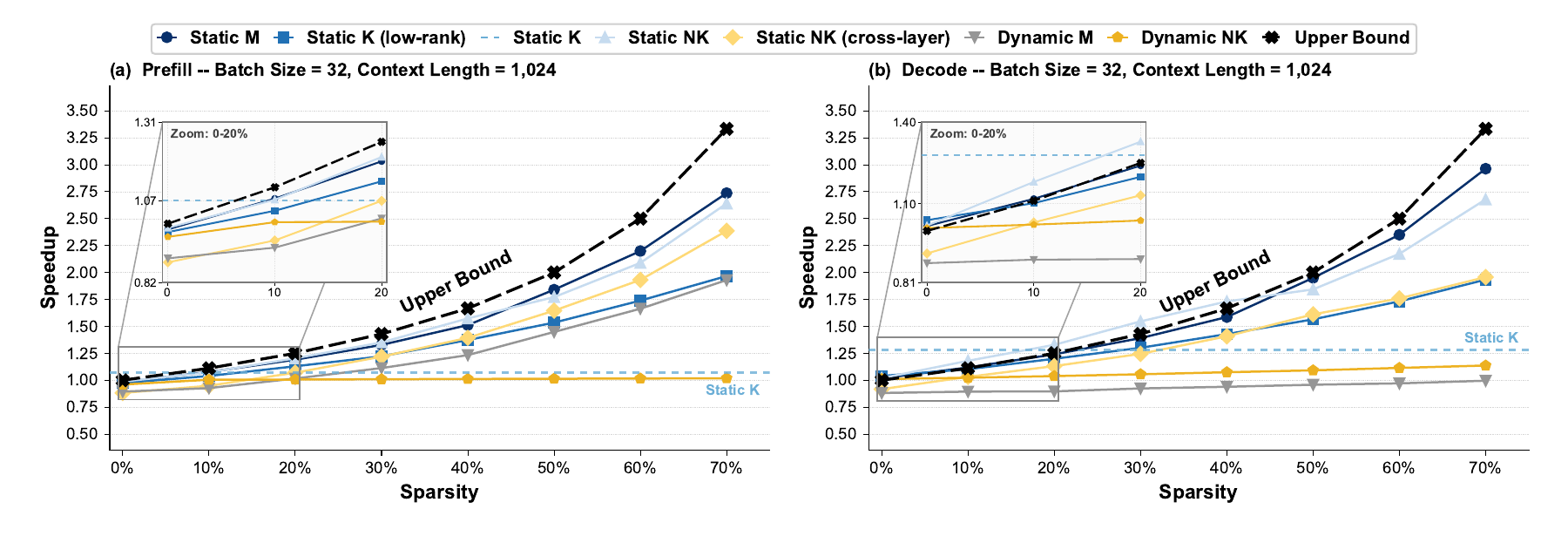}
  }
  \caption{Speedup over different sparsity, and their gap with the theoretical upper bound. Context length is set to 1,024.}
  \label{fig:sm120_50_appendix}
  \vspace{-5pt}
\end{figure*}

\label{sec:appendix}

\end{document}